\documentclass[11pt]{article}

\usepackage[final]{acl}

\usepackage{times}
\usepackage{latexsym}

\usepackage[T1]{fontenc}
\usepackage[utf8]{inputenc}

\usepackage{microtype}

\usepackage{inconsolata}

\usepackage{graphicx}
\usepackage{xcolor}
\usepackage{colortbl}
\usepackage{booktabs}
\usepackage{amsmath}
\usepackage{multirow}
\usepackage{pifont}
\usepackage[most]{tcolorbox}
\graphicspath{{figures/}}

\newcommand{\method}{\textsc{MemWM}}
\newcommand{\circnum}[1]{\ding{\numexpr181+#1\relax}}
\definecolor{questioncream}{HTML}{FDF8EA}
\definecolor{algpurple}{HTML}{7A3FC8}
\definecolor{algcomment}{HTML}{9B84D8}
\definecolor{alfcol}{HTML}{F1FAFF}
\definecolor{webcol}{HTML}{FFF6EE}
\definecolor{scicol}{HTML}{F3FAEF}
\definecolor{promptolive}{HTML}{536347}
\definecolor{promptcream}{HTML}{FDF8EA}
\newtcolorbox{questionbox}{
  colback=questioncream,
  colframe=gray!55,
  boxrule=0.6pt,
  arc=3pt,
  left=6pt,
  right=6pt,
  top=5pt,
  bottom=5pt,
  width=0.92\linewidth,
  center,
  before skip=0.5\baselineskip,
  after skip=0.5\baselineskip
}
\newtcblisting{promptbox}[1][]{%
  colback=promptcream,
  colframe=promptolive,
  colbacktitle=promptolive,
  coltitle=white,
  fonttitle=\small\bfseries,
  boxrule=0.7pt,
  arc=2.5mm,
  left=6pt,
  right=6pt,
  top=6pt,
  bottom=6pt,
  breakable,
  listing only,
  listing options={basicstyle=\ttfamily\footnotesize,breaklines=true,columns=fullflexible,keepspaces=true},
  title=#1
}

\title{%
  \texorpdfstring{%
    \makebox[\textwidth][c]{%
      \raisebox{-0.38\height}{%
        \includegraphics[height=1.8cm]{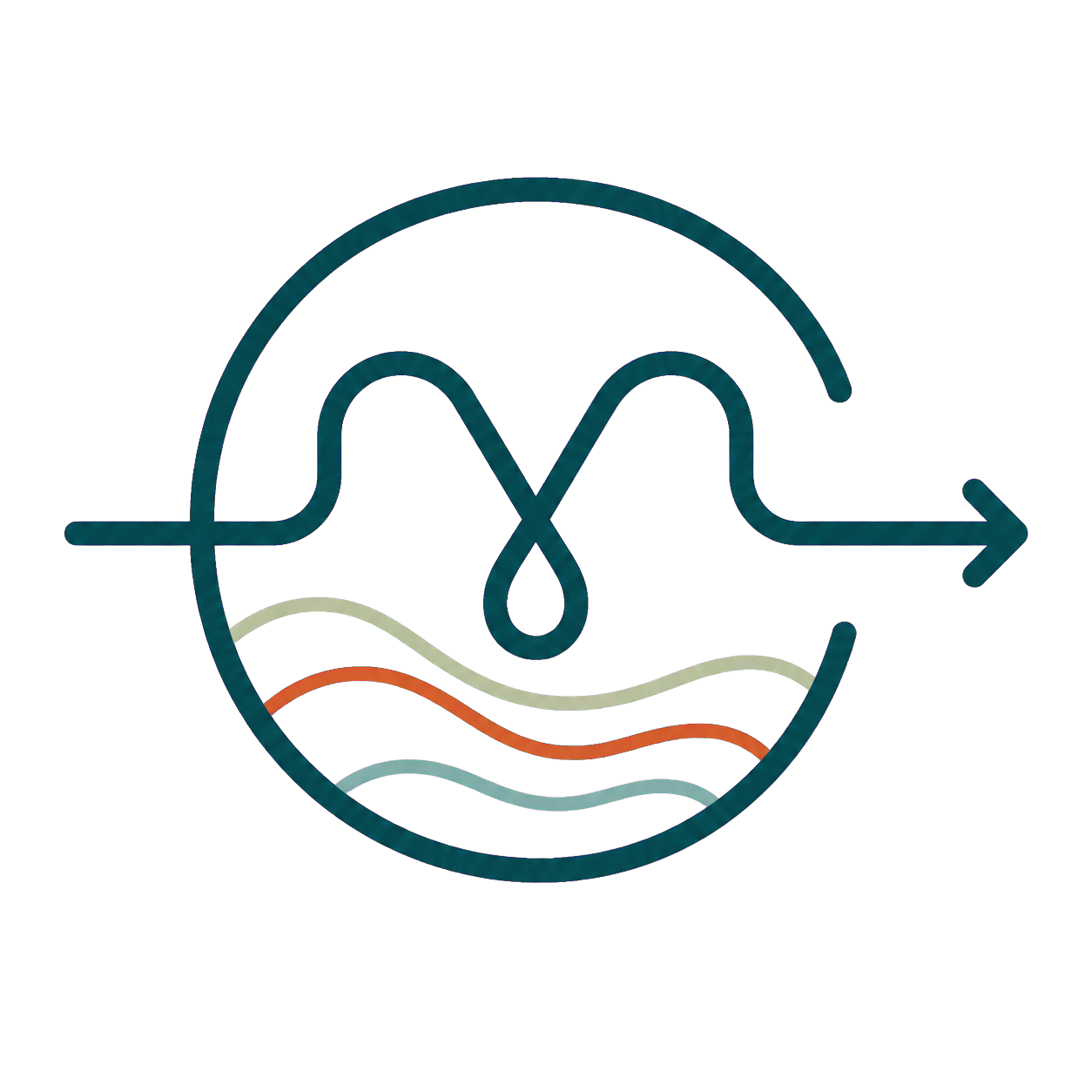}%
      }%
      \hspace{0.6em}%
      \parbox[c]{0.78\textwidth}{%
        \centering
        \method: Memory-Augmented Text-Based World Model
      }%
    }%
  }{%
    MemWM: Memory-Augmented Text-Based World Model
  }%
}

\author{
  \textbf{Yujun Wang}%
  \textsuperscript{1,2,3 *}
  \quad
  \textbf{Tao Zhang}%
  \textsuperscript{4 *}
  \quad
  \textbf{Jinhe Bi}%
  \textsuperscript{1,2}
  \quad
  \textbf{Aniri}%
  \textsuperscript{1} \\[-0.1em]
  \textbf{Wenxuan Ye}%
  \textsuperscript{3,5}
  \quad
  \textbf{Boliang Liu}%
  \textsuperscript{3,6}
  \quad
  \textbf{Sikuan Yan}%
  \textsuperscript{1,2,3}
  \quad
  \textbf{Shuning Wang} \\[-0.1em]
  \textbf{Xuebing Zhou}%
  \textsuperscript{3}
  \quad
  \textbf{S{\"o}ren Pirk}%
  \textsuperscript{7}
  \quad
  \textbf{Hinrich Sch{\"u}tze}%
  \textsuperscript{1,2}
  \quad
  \textbf{Yunpu Ma}%
  \textsuperscript{1,2}
}

\begin{document}
\maketitle

% \begin{NoHyper}
% \begingroup
% \renewcommand{\thefootnote}{}
% \footnotetext{%
%     \textbf{Accepted to EMNLP 2026 (Main Conference).}
%     \par\smallskip

%   \textsuperscript{1}LMU Munich;
%   \textsuperscript{2}Munich Center for Machine Learning (MCML);
%   \textsuperscript{3}Huawei Heisenberg Research Center;
%   \textsuperscript{4}Zhejiang University;
%   \textsuperscript{5}Technical University of Munich (TUM);
%   \textsuperscript{6}TU Berlin;
%   \textsuperscript{7}Kiel University.
%   \quad
%   \textsuperscript{*}Equal contribution.
%   \quad
%   Corresponding authors:
%   Yujun Wang (\texttt{yujun\_wang\_cn@hotmail.com}) and
%   Yunpu Ma (\texttt{cognitive.yunpu@gmail.com}).
% }
% \endgroup
% \end{NoHyper}

\begin{NoHyper}
\makeatletter
\begingroup
\renewcommand{\thefootnote}{}
\renewcommand{\@makefntext}[1]{\noindent #1}

\footnotetext{%
\raggedright
\textbf{Accepted to EMNLP 2026 (Main Conference).}
\par\smallskip

\textsuperscript{1}\,LMU Munich;
\textsuperscript{2}\,Munich Center for Machine Learning (MCML);
  \textsuperscript{3}Huawei Heisenberg Research Center;
  \textsuperscript{4}Zhejiang University;
  \textsuperscript{5}Technical University of Munich (TUM);
  \textsuperscript{6}TU Berlin;
  \textsuperscript{7}Kiel University.
\par\smallskip

\textsuperscript{*}\,Equal contribution.
\quad
Corresponding authors:
Yujun Wang (\texttt{yujun\_wang\_cn@hotmail.com}) and
Yunpu Ma (\texttt{cognitive.yunpu@gmail.com}).
}

\endgroup
\makeatother
\end{NoHyper}

\setcounter{footnote}{0}

\begin{abstract}
World models are increasingly used to support planning in agents by predicting how environment states evolve in response to agent actions. Yet fluent next-state predictions can still omit task-critical facts, corrupt product attributes, or apply incorrect transition rules. To address such systematic prediction errors, we introduce \method{}, a memory-augmented text-based world model. \method{} uses \emph{world memory}, a curated memory bank of transition rules, state caches, and hard-to-predict facts, to condition next-state imagination. We evaluate factual state preservation with Structured State Fidelity (SSF), which scores predicted states through benchmark-specific facts and fields. Compared with SFT, memory-augmented training improves SSF by up to 206.3\%. In the full planning setting, we keep the policy model frozen and provide policy-side \emph{world skill}: retrieved task-level skills and step-wise corrective guidance for action selection. Across ALFWorld, WebShop, and ScienceWorld, memory-augmented agents improve downstream success over an SFT-trained world-model agent, with up to a 65.4\% relative gain. Sensitivity analyses further show that retrieved memory improves task success and efficiency under different memory and action-budget settings.
\end{abstract}

\section{Introduction}

\begin{figure*}[t]
  \centering
  \includegraphics[width=\textwidth]{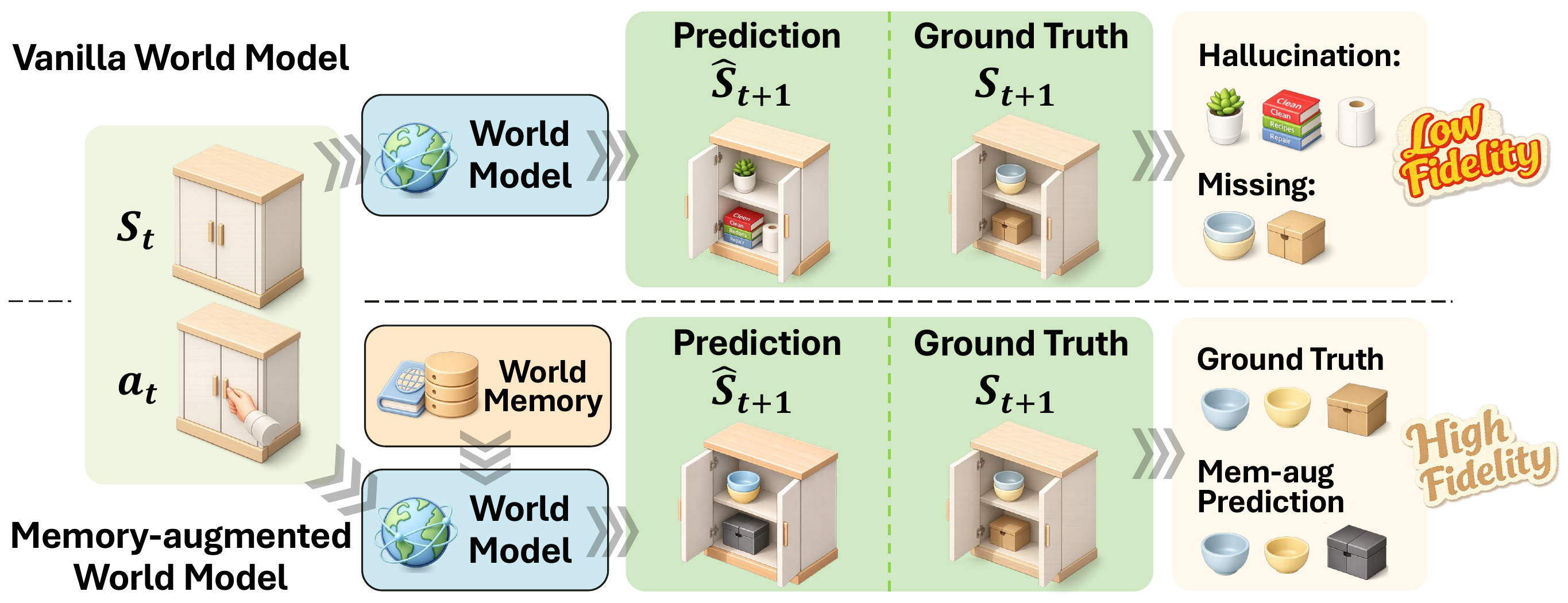}
  \caption{
  \textbf{Illustration of the state-fidelity bottleneck in world modeling.}
  \textbf{Upper:} Given the current state $s_t$ and action $a_t$, a world model without retrieved memory may produce a fluent but low-fidelity next state, hallucinating objects, missing task-relevant facts, or applying the wrong transition rule.
  \textbf{Lower:} By retrieving curated world memory, \method{} conditions prediction on relevant state caches, transition rules, and hard-to-predict facts, producing an imagined state that better matches the ground truth $s_{t+1}$.
  }
  \label{fig:teaser}
\end{figure*}

Large language model agents are increasingly used for sequential decision making in interactive environments, ranging from embodied household tasks to scientific simulations and web navigation \citep{yao2023react,ahn2022saycan,park2023generativeagents,wang2023voyager,shinn2023reflexion}. A central ingredient behind such agents is the ability to reason about how the world changes after an action. World models provide this capability by allowing an agent to imagine future states, compare candidate trajectories, and plan before acting \citep{sutton1991dyna,ha2018worldmodels,hafner2019dream,janner2019mbpo,schrittwieser2020muzero}. Recent work has therefore begun to study language models as implicit or explicit text-based world models for planning and decision making \citep{qiao2024wkm,li2025wordtoworld,liu2026itp,huang2026behaviorconsistency}.

Despite this promise, text-based world models face a state-fidelity bottleneck. A predicted next state can be fluent and locally plausible while still losing the facts that determine whether a future action is valid or useful. For example, an embodied agent may remember that it moved to a room but forget which object was left in which receptacle; a scientific environment may preserve the wording of an observation while changing a device state or numeric measurement; a shopping agent may keep the user's broad intent while corrupting a product identifier, price, or option. Such errors are easy to overlook at the surface level, but they are precisely the errors that compound during lookahead planning.

This bottleneck is partly hidden by standard next-state prediction metrics. Exact match is too strict: harmless paraphrases, formatting changes, or reordered facts can receive zero credit. Word F1 is too permissive: predictions that swap an object, location, state, product ID, or price can still share most of their tokens with the gold state. As a result, a world model can appear strong under surface overlap while remaining unreliable as a planning component. This motivates evaluating world models by whether they preserve structured, behavior-relevant state facts rather than by whether they reproduce the same string.

Figure~\ref{fig:teaser} illustrates this motivation: memory is useful not because it makes the prediction longer, but because it helps preserve the structured state facts that determine whether the imagined future is faithful.

This raises the central question of this paper:
\begin{questionbox}
\emph{How can a text-based world model use experience-derived memory to avoid systematic fact, rule, and state-transition errors while keeping imagined states faithful to behavior-critical world facts?}
\end{questionbox}

In response to this question, we propose \method, a memory-augmented text-based world model for state-faithful imagination. The core idea is to equip the world model with \emph{world memory}, a curated memory bank of transition rules, state caches, and hard-to-predict facts that help next-state prediction avoid recurring errors. In the full agent, \method{} is further coupled with policy-side \emph{world skill}, providing task-level skills and step-wise corrective guidance that strengthen action selection.

We evaluate this idea across three text-based decision-making benchmarks: ALFWorld, ScienceWorld, and WebShop. These benchmarks stress different forms of state fidelity: object locations and receptacle states in embodied household interaction, scientific facts and numeric/device states in ScienceWorld, and product identities, prices, options, and user constraints in WebShop. This diversity allows us to test whether memory augmentation improves world modeling beyond a single environment template.

Our contributions are organized around three points:
\begin{itemize}
    \item[\circnum{1}] \textbf{Structured state-fidelity evaluation.} We show that surface next-state metrics can obscure behavior-critical fact, rule, and state-transition errors, and introduce Structured State Fidelity (SSF) to evaluate predicted states by comparing benchmark-specific world facts rather than surface string overlap.
    \item[\circnum{2}] \textbf{World-memory-augmented world modeling.} We introduce a memory-augmented world model that retrieves curated world memory, including transition rules, state caches, and hard-to-predict facts, to improve the fidelity of imagined next states.
    \item[\circnum{3}] \textbf{Stronger agents without policy training.} With frozen policy models, memory-augmented world models improve downstream success across ALFWorld, ScienceWorld, and WebShop; policy-side world skill further strengthens action selection by adding task-level skills and step-wise corrective guidance. Sensitivity analyses further show that these gains are not merely due to larger action budgets and can improve successful-task efficiency.
\end{itemize}

\section{Related Work}

\begin{figure*}[t]
  \centering
  \includegraphics[width=\textwidth]{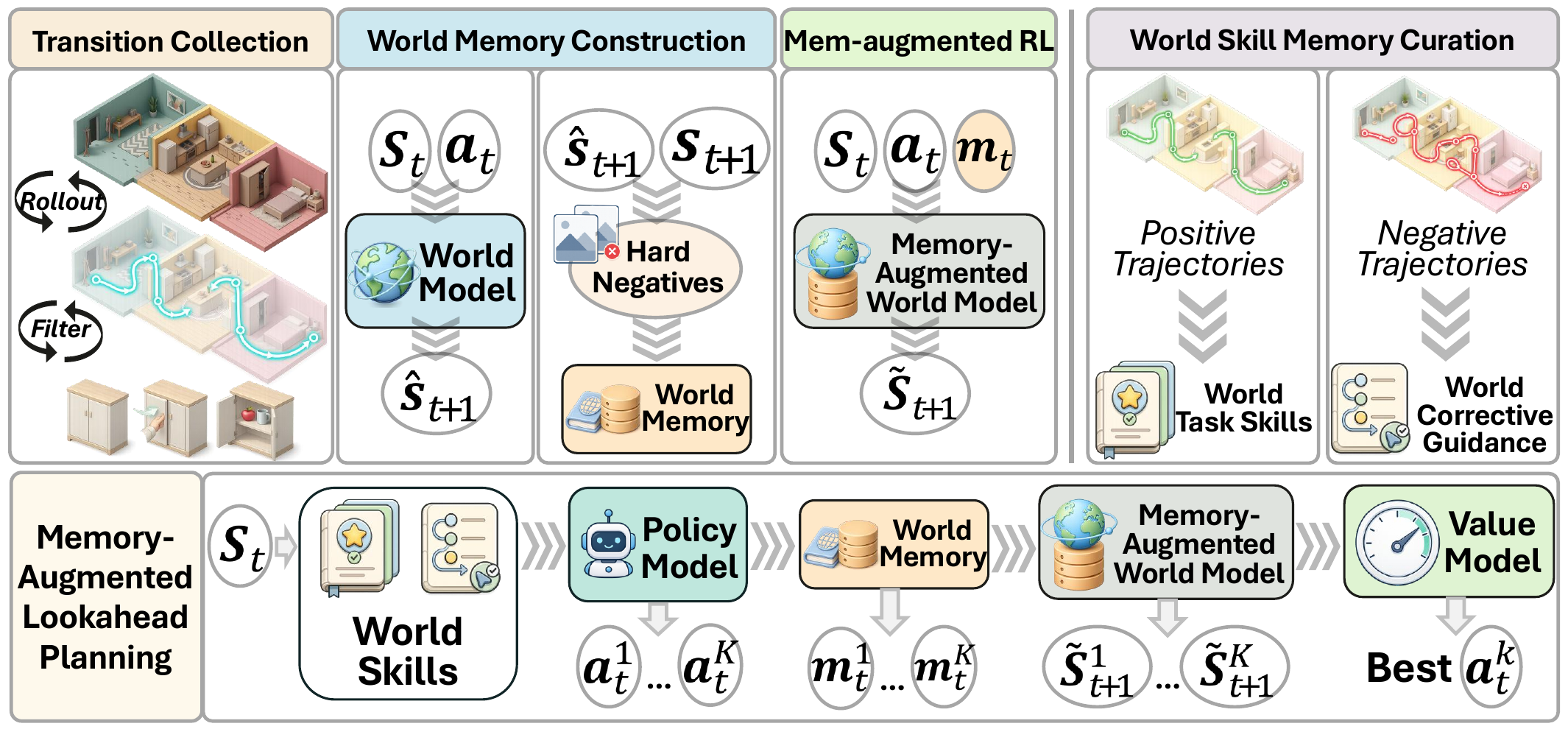}
  \caption{
  \textbf{Overview of the \method{} planning pipeline.}
  \textbf{Top:} The left branch builds reusable state and transition memories; the right branch builds reusable action-selection experience.
  \textbf{Bottom:} At inference time, the agent retrieves relevant entries, imagines candidate next states, scores the candidates, and commits to the highest-scoring action in the environment.
  }
  \label{fig:main-method}
\end{figure*}

\paragraph{Text-based world models for agents.}
Text-based games and interactive language environments have long served as testbeds for agents that infer latent state, track objects, and act through natural-language commands \citep{narasimhan2015textgames,cote2018textworld,hausknecht2020jericho,shridhar2021alfworld,wang2022scienceworld,yao2022webshop}. Recent work studies language models as explicit or implicit text-based world models: world-knowledge models inject environment knowledge \citep{qiao2024wkm}, Word-to-World analyzes internal text-world dynamics \citep{li2025wordtoworld}, and lookahead planners use imagined trajectories for action selection \citep{liu2026itp}. These directions show that planning can benefit from imagined states, but remains sensitive to fine-grained state errors.

\paragraph{Evaluating state fidelity.}
Automatic text-generation metrics usually measure surface overlap or semantic similarity, as in BLEU, ROUGE, and BERTScore \citep{papineni2002bleu,lin2004rouge,zhang2020bertscore}. Factuality metrics decompose text into finer-grained units \citep{min2023factscore}, and recent text-world evaluation emphasizes downstream behavioral consistency \citep{huang2026behaviorconsistency}. Complementary decoding-time work addresses hallucination at generation time: ASCD steers cross-modal attention to reduce hallucination in multimodal language models \citep{wang2026ascd}. For text-based world models, this distinction matters because paraphrases and behavior-critical corruptions can receive misleading overlap scores. SSF follows this fact-oriented view with benchmark-specific state structure.

\paragraph{Memory-augmented language systems.}
Retrieval-augmented generation and dense retrieval give language models access to external evidence beyond parametric memory \citep{guu2020realm,karpukhin2020dpr,lewis2020rag,izacard2021fid}. Long-term memory systems further study how stored context can support recall, consistency, and adaptation over time \citep{borgeaud2021retro,wang2023longmem,zhong2023memorybank}. In contrast to general knowledge retrieval, our memory is organized around world-model prediction errors, transition patterns, and reusable state facts.

\paragraph{Policy-side skills and feedback.}
A separate line of agent work improves action selection through reusable skills or feedback from past trials. Reflexion stores verbal feedback for future attempts \citep{shinn2023reflexion}, Voyager builds a skill library during open-ended exploration \citep{wang2023voyager}, and SkillRL studies recursive skill-augmented reinforcement learning \citep{xia2026skillrl}. We use policy-side world skill only as retrieval-time guidance for the frozen policy in the full agent.

\paragraph{Broader LLM reasoning and training.}
Complementary work investigates rollout reuse and reinforcement mid-training for LLM reasoning \citep{bi2026echorl,mid}, layerwise trajectory self-evaluation and verifier-guided long-chain-of-thought synthesis \citep{bi2026the,huang2025loongsynthesizelongchainofthoughts}, and the broader landscape of multimodal code intelligence \citep{zhao2026nl2codestructuredsurveymultimodal}. These directions address general reasoning, training, and code-oriented capabilities, whereas our focus is state-faithful prediction and retrieval-time memory in interactive text environments.

\section{Preliminaries}

This section defines the next-state prediction setting and Structured State Fidelity (SSF), which we use to evaluate whether imagined states preserve behavior-critical world facts.

\subsection{Task Formulation and Notation}

We consider a language-agent interaction over discrete time steps. At step $t$, the agent observes a textual state $s_t$, takes an action $a_t \in \mathcal{A}$, and receives the next textual state $s_{t+1}$. We write the interaction history as $h_t = (s_1, a_1, \ldots, s_t)$, denote world memory by $\mathcal{M}^{\mathrm{wm}}_t$, and denote policy-side world skill by $\mathcal{M}^{\mathrm{skill}}_t$. A text-based world model $f_\theta$ predicts the consequence of a candidate action by generating
\begin{equation}
    \hat{s}_{t+1} = f_\theta(c_t, a_t),
\end{equation}
where $c_t$ denotes the context available to the model. In the common local-context setting, $c_t$ contains only the current observation or a short recent trajectory window.

The world-model-side memory $\mathcal{M}^{\mathrm{wm}}_t$, or \emph{world memory}, stores persistent task information extracted from $h_t$. A memory-augmented world model predicts
\begin{equation}
    \hat{s}_{t+1}
    = f_\theta\left(c_t, a_t,
    R_{\mathrm{wm}}(q^{\mathrm{wm}}_t, \mathcal{M}^{\mathrm{wm}}_t)\right),
\end{equation}
where $q^{\mathrm{wm}}_t$ is a retrieval query derived from the current context and candidate action, and $R_{\mathrm{wm}}$ returns a compact set of relevant world memory entries. These entries provide state facts, transition rules, or constraints that help the model predict the action-conditioned next state more faithfully.

For policy-side behavior, we use a separate memory channel $\mathcal{M}^{\mathrm{skill}}_t$, or \emph{world skill}, to condition the acting policy on reusable experience. World skill has two forms: \emph{world task skill}, a task-level prior retrieved once at the beginning of an episode, and \emph{world corrective guidance}, step-level guidance retrieved at each decision step.

\subsection{Structured State Fidelity}
\label{sec:ssf}

Existing text metrics such as exact match and word-level F1, in the same family of surface-overlap evaluation as BLEU and ROUGE \citep{papineni2002bleu,lin2004rouge}, are poorly aligned with text-based world modeling. Exact match is too strict because harmless paraphrases, formatting differences, and ordering changes can make a correct state receive zero credit. Word F1 is often too permissive because a prediction can share most tokens with the gold state while corrupting the facts that determine future behavior, such as swapping an object identity, changing a receptacle, flipping an open/closed state, or replacing a product price.

\begin{figure}[t]
  \centering
  \includegraphics[width=\columnwidth]{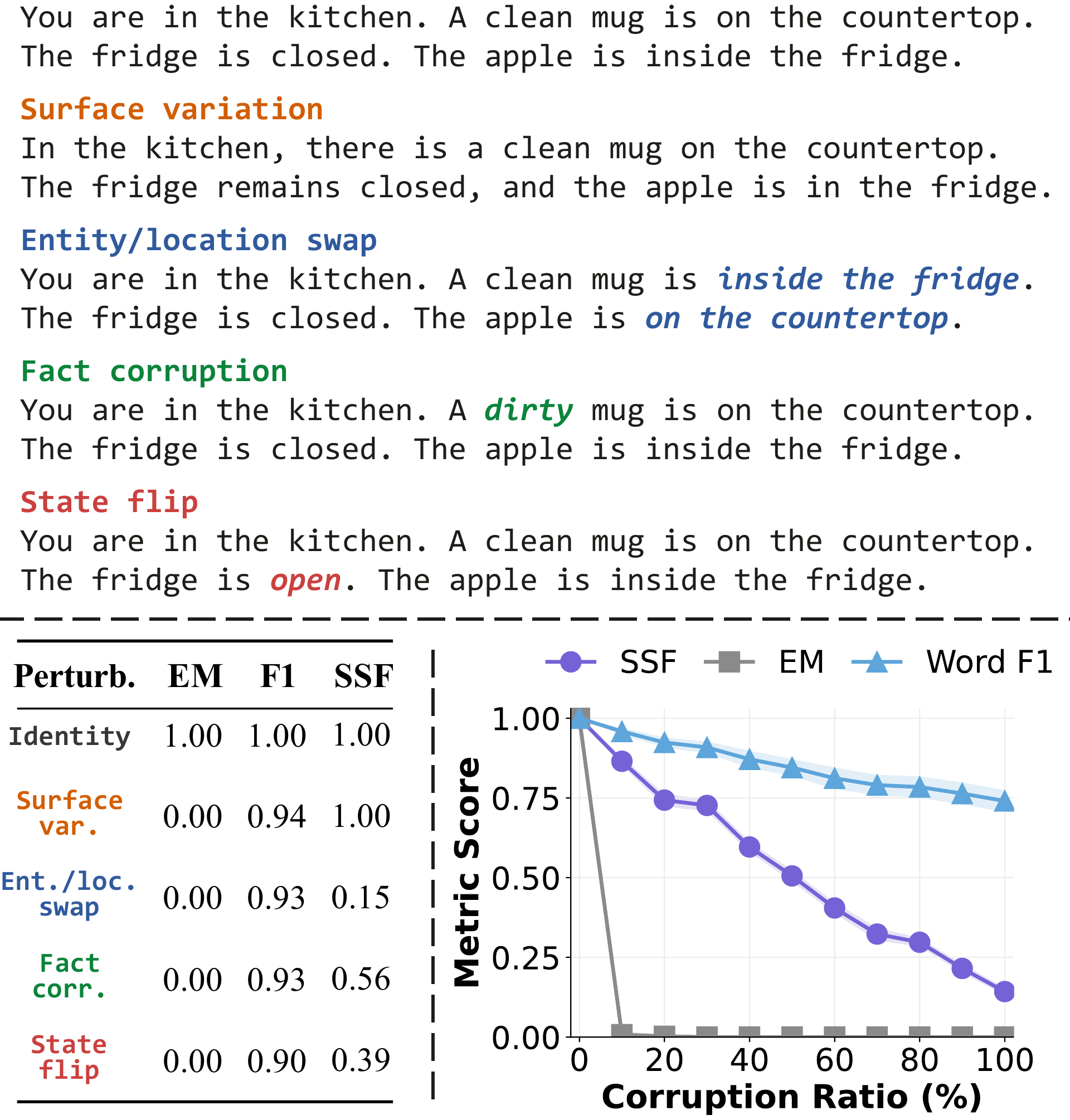}
  \caption{
  \textbf{Metric stress test motivating Structured State Fidelity.}
  Exact match is brittle under non-identical but faithful states, while Word F1 can remain high after behavior-critical fields are corrupted. SSF more directly tracks degradation in structured state information.
  }
  \label{fig:ssf-motivation}
\end{figure}

Figure~\ref{fig:ssf-motivation} visualizes this mismatch with controlled perturbations of gold next states. To address it, we propose \textbf{Structured State Fidelity} (\textbf{SSF}), a domain-aware metric that compares predicted and gold states after projecting them into structured world facts. For a domain $d$, let $\Phi_d$ be a state extractor, and let $\mathcal{C}_d$ denote the set of structured components used in that domain. For each component $c$, let $\hat{z}_c=\Phi_{d,c}(\hat{s}_{t+1})$ and $z_c=\Phi_{d,c}(s_{t+1})$ denote the predicted and gold structured values. SSF is defined as
\begin{equation}
    \mathrm{SSF}_d(\hat{s}_{t+1}, s_{t+1})
    = \sum_{c \in \mathcal{C}_d} w_c \cdot
    \mathrm{sim}_c(\hat{z}_c, z_c),
    \label{eq:ssf-general}
\end{equation}
where $w_c$ is a component weight and $\mathrm{sim}_c$ compares the predicted and gold values of component $c$. Depending on the component type, $\mathrm{sim}_c$ can be exact match for categorical fields, set F1 for fact or identifier sets, lexical F1 for short text fields, or a relative-error score for numeric fields such as prices. This formulation makes the metric explicit about which state fields matter, while allowing the extractor and component set to match the structure of each environment.

Algorithm~1 summarizes how SSF is computed in our experiments. ALFWorld and ScienceWorld share the same fact-set view: both predicted and gold states are projected into normalized canonical facts, and SSF is their fact-level F1. WebShop follows the same principle but uses page-specific structured fields, because its observations are search-result pages, product pages, or terminal pages rather than generic fact lists.

\begin{center}
\begin{minipage}{\columnwidth}
  \small
  \setlength{\tabcolsep}{0pt}
  \renewcommand{\arraystretch}{1.08}
  \begin{tabular}{@{}r@{\hspace{0.35em}}p{0.82\columnwidth}@{}}
    \toprule
    \multicolumn{2}{@{}l}{\textbf{Algorithm 1 Structured State Fidelity (SSF)}} \\
    \midrule
    1: & \textbf{Input:} $\hat{s}_{t+1}$, $s_{t+1}$, domain $d$. \textbf{Output:} $\mathrm{SSF}_d$. \\
    2: & \textcolor{algpurple}{\textbf{Fact-Set SSF:}} \\
    3: & \textbf{if} $d \in \{\textcolor{algpurple}{\textsc{ALFWorld}}, \textcolor{algpurple}{\textsc{ScienceWorld}}\}$ \textbf{then} \\
    4: & \quad Instantiate fact extractor $\Phi_d$ with a tiny LLM. \\
    5: & \quad $\hat{F},F\leftarrow\Phi_d(\hat{s}_{t+1}),\Phi_d(s_{t+1})$. \\
    6: & \quad \textbf{return} fact-F1$(\hat{F}, F)$. \textcolor{algcomment}{\(\triangleright\) fact-set score} \\
    7: & \textcolor{algpurple}{\textbf{Page-Field SSF:}} \\
    8: & \textbf{else if} $d=\textcolor{algpurple}{\textsc{WebShop}}$ \textbf{then} \\
    9: & \quad Instantiate extractor $\Phi_d$ with a rule-based page parser. \\
    10: & \quad $\hat{z},z\leftarrow\Phi_d(\hat{s}_{t+1}),\Phi_d(s_{t+1})$. \\
    11: & \quad Compute $\mathrm{sim}_c(\hat{z}_c,z_c)$ for each field $c$. \\
    12: & \quad \textbf{return} weighted score. \textcolor{algcomment}{\(\triangleright\) field score} \\
    13: & \textbf{end if} \\
    \bottomrule
  \end{tabular}
\end{minipage}
\end{center}

SSF is designed to be a metric, not an unrestricted LLM judge. Whenever a domain has stable textual templates, we prefer deterministic extraction rules because they are transparent, reproducible, and less prone to completing missing information. Appendix~\ref{sec:appendix-ssf} gives the benchmark-specific fact schemas, WebShop fields, parser details, and sanity checks.

\section{\method}

\subsection{Overview}

\method{} is motivated by recurring world-model errors in facts, rules, and state transitions. A text-based world model may generate a fluent next state while omitting an object binding, corrupting a product attribute, or applying an incorrect environment rule. To reduce such errors, \method{} augments next-state prediction with \emph{world memory}, whose entries are derived from previous observations, actions, inferred state facts, and curated transition knowledge. SSF then provides a structured way to measure whether this memory improves state fidelity.

World memory conditions next-state imagination, while a separate policy-side world skill channel can be used in the full agent to condition action selection. The two channels are used at different points in planning: one supports consequence prediction, and the other supports deciding which actions to consider.

Algorithm~2 summarizes how these components are used during planning. The algorithm is independent of the specific planner: the scoring rule can be greedy action selection, lookahead search, or another downstream decision rule.

\begin{center}
\begin{minipage}{\columnwidth}
  \small
  \setlength{\tabcolsep}{0pt}
  \renewcommand{\arraystretch}{1.08}
  \begin{tabular}{@{}r@{\hspace{0.35em}}p{0.82\columnwidth}@{}}
    \toprule
    \multicolumn{2}{@{}l}{\textbf{Algorithm 2 \method{} Planning}} \\
    \midrule
    1: & \textbf{Input:} goal $g$, state $s_1$, world memory $\mathcal{M}^{\mathrm{wm}}$, skill memory $\mathcal{M}^{\mathrm{skill}}$. \\
    2: & Initialize history $h_1=(s_1)$. \\
    3: & $\tau \leftarrow R_{\mathrm{task}}(g,\mathcal{M}^{\mathrm{skill}})$. \\
    4: & \textbf{for} decision step $t=1,\ldots,T$ \textbf{do} \\
    5: & \quad Build world-model context $c_t$ from $h_t$ and $s_t$. \\
    6: & \quad $\gamma_t \leftarrow R_{\mathrm{corr}}(c_t,\mathcal{M}^{\mathrm{skill}})$. \\
    7: & \quad Propose candidate actions $\mathcal{A}_t \leftarrow \pi_\psi(c_t,\tau,\gamma_t)$. \\
    8: & \quad \textcolor{algpurple}{\textbf{Memory-Conditioned Imagination:}} \\
    9: & \quad \textbf{for each} $a \in \mathcal{A}_t$ \textbf{do} \\
    10: & \quad\quad $m_t(a)\leftarrow R_{\mathrm{wm}}(c_t,a,\mathcal{M}^{\mathrm{wm}})$. \\
    11: & \quad\quad $\hat{s}_{t+1}(a)\leftarrow f_\theta(c_t,a,m_t(a))$. \\
    12: & \quad\quad $J(a)\leftarrow \mathrm{Score}(a,\hat{s}_{t+1}(a),h_t)$. \\
    13: & \quad \textbf{end for} \\
    14: & \quad Execute $a_t\leftarrow\arg\max_{a\in\mathcal{A}_t}J(a)$; observe $s_{t+1}$. \\
    15: & \quad Update $h_{t+1}$ and memory evidence. \\
    16: & \textbf{end for} \\
    \bottomrule
  \end{tabular}
\end{minipage}
\end{center}

\subsection{Memory Construction}

\paragraph{World memory.}
The world memory module stores compact keyed entries distilled from past world-model experience, structured states, and recurring transition patterns. Each entry is curated as a transition rule, a state-cache record, or a hard-to-predict fact that the model may otherwise omit or corrupt. In household tasks, entries can encode action preconditions, object-state changes, and receptacle relations; in ScienceWorld, experimental preconditions, recipes, temperatures, and progress markers; in WebShop, product identifiers, prices, option values, and user constraints. Entries are kept concise and keyed for lightweight retrieval.

\paragraph{World skill.}
The policy-side memory stores reusable behavior-level experience. A \emph{world task skill} summarizes a high-level strategy for a task family, such as which subgoals to pursue first or which page fields to inspect before buying a product. A \emph{world corrective guidance} entry records a local action correction, such as avoiding an invalid repeated action, revisiting a needed object, checking an option before purchasing, or preserving a scientific precondition before the next manipulation. These entries are phrased as guidance for the policy, not as state facts for the world model.

\subsection{Memory Retrieval}

Given a state-prediction prompt, \method{} retrieves a compact set of world-memory entries using structured cues from the current context and the action being predicted. Retrieval is instantiated with lightweight domain-specific matching, so the returned entries are tied to the predicted transition rather than generic background knowledge. If retrieved memory conflicts with the visible trajectory, the trajectory remains authoritative.

World skill follows a separate retrieval schedule. Task-level skill is retrieved once from the task instruction or goal, while corrective guidance is retrieved at each decision step from the current observation, recent action history, and available failure or planner signals. This schedule keeps policy advice separate from world-state memory while allowing both channels to inform planning.

\subsection{Memory-Conditioned Imagination}

The world model predicts the next state from the local trajectory context plus the retrieved world-memory entries. Rather than changing the decoding objective, \method{} inserts the retrieved entries as auxiliary evidence in the state-prediction prompt, making relevant constraints visible at generation time. The model then generates a fluent next state conditioned on both the immediate observation-action context and the retrieved evidence.

\subsection{Integration with Planning}

\method{} separates action selection from consequence prediction while allowing both to benefit from memory. At task start, the policy retrieves a world task skill to form a high-level solving prior. At each decision step, world corrective guidance helps the policy propose candidate actions. The planner then evaluates these candidates by querying the memory-augmented world model: for each candidate, world memory is retrieved, the next state is imagined, and the resulting trajectory is scored by the downstream planning rule. After execution, the observed transition can be added back to the memory bank, allowing later predictions and decisions to reuse the new evidence.

\begin{table}[!t]
  \centering
  \begingroup
  \setlength{\tabcolsep}{3.0pt}
  \renewcommand{\arraystretch}{1.08}
  \begin{tabular}{clccc}
    \toprule
    \textbf{Model} & \textbf{Method} & \textbf{ALF.} & \textbf{Web.} & \textbf{Sci.} \\
    \midrule
    \multirow{3}{*}{Llama3.2-1B} & SFT & 0.815 & 0.277 & 0.887 \\
    & RL & 0.835 & 0.464 & 0.895 \\
    & Mem-Aug RL & \textbf{0.865} & \textbf{0.639} & \textbf{0.912} \\
    \midrule
    \multirow{3}{*}{Qwen3-4B} & SFT & 0.855 & 0.726 & 0.884 \\
    & RL & 0.861 & 0.814 & 0.902 \\
    & Mem-Aug RL & \textbf{0.894} & \textbf{0.884} & \textbf{0.915} \\
    \midrule
    \multirow{3}{*}{Qwen2.5-7B} & SFT & 0.286 & 0.561 & 0.880 \\
    & RL & 0.848 & 0.664 & 0.903 \\
    & Mem-Aug RL & \textbf{0.876} & \textbf{0.799} & \textbf{0.915} \\
    \bottomrule
  \end{tabular}
  \endgroup
  \caption{
  \textbf{Structured State Fidelity (SSF) of world-model predictions.}
  Scores measure preservation of benchmark-specific structured state facts. ALF., Web., and Sci. denote ALFWorld, WebShop, and ScienceWorld, respectively. RL denotes reinforcement learning without memory; Mem-Aug RL denotes memory-augmented reinforcement learning. Best scores within each model block and benchmark are bolded.
  }
  \label{tab:ssf-results}
\end{table}

\section{Experiment}

In this section, we evaluate \method{} through three questions: \textbf{RQ1}: does \emph{world memory} improve structured next-state fidelity? \textbf{RQ2}: do \emph{world memory} and policy-side \emph{world skill} improve downstream agent performance? \textbf{RQ3}: how sensitive are the gains to memory availability and the action budget?

\subsection{Experiment Setup}

\paragraph{Benchmarks.}
We evaluate on \textbf{ALFWorld} \citep{shridhar2021alfworld}, \textbf{ScienceWorld} \citep{wang2022scienceworld}, and \textbf{WebShop} \citep{yao2022webshop}. These benchmarks test complementary forms of structured state fidelity: household object states and receptacles, scientific entities and procedural preconditions, and shopping-page fields such as product identifiers, prices, options, and user constraints.

\begin{figure}[!t]
  \centering
  \includegraphics[width=\columnwidth]{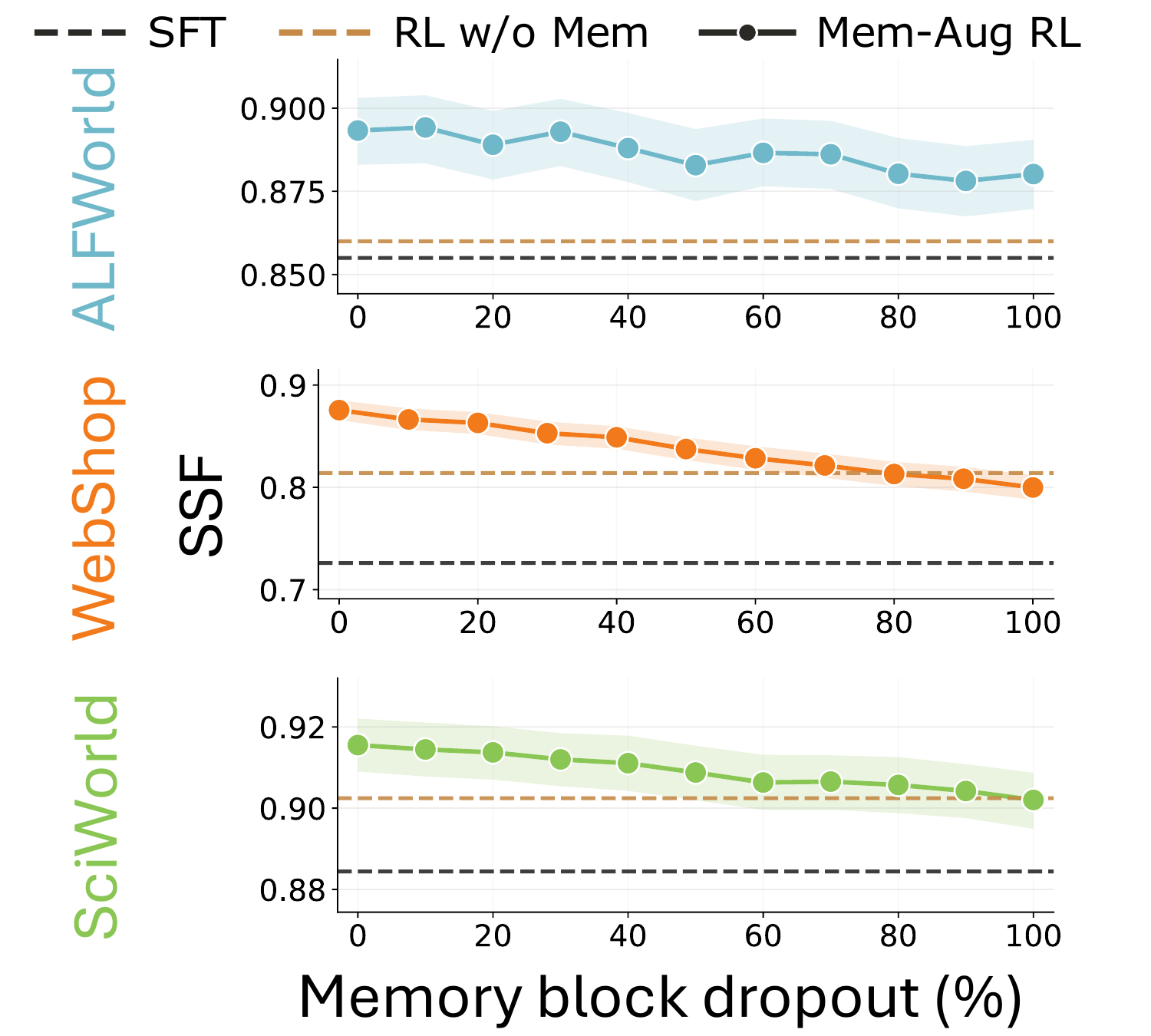}
  \caption{
  \textbf{SSF sensitivity to world-memory dropout.}
  With Qwen3-4B, increasing prediction-time memory dropout reduces SSF across ALFWorld, WebShop, and ScienceWorld.
  }
  \label{fig:memory-dropout}
\end{figure}

\paragraph{Metrics.}
For next-state prediction, we report Structured State Fidelity (SSF), since exact match and word-level F1 do not capture the behavior-critical facts that matter for text-based world models in our setting. We use exact match and word F1 only in the metric stress test that motivates SSF. For downstream behavior, we report task success; in sensitivity analyses, we also report successful-task average steps when relevant.

\paragraph{Baselines.}
For world-model training, we compare SFT, RL without memory, and memory-augmented RL. For downstream agent evaluation, we compare against CoT, ReAct, RAP, ITP where applicable, an SFT-trained world-model (WM) agent, and memory-augmented variants with and without policy-side world skill.

\paragraph{Implementation.}
All methods use the same benchmark interface, action space, decoding setup, and evaluation budget within each benchmark. In our downstream agent rows, the policy backbone is kept frozen: we do not fine-tune it with supervised learning or reinforcement learning. The differences come from the world model, retrieved world memory, and optional retrieval-time world skill. Benchmark and implementation details are provided in Appendices~\ref{sec:appendix-benchmark-setup} and~\ref{sec:appendix-implementation}.

\begin{table*}[!t]
  \centering
  \begingroup
  \setlength{\tabcolsep}{1.5pt}
  \renewcommand{\arraystretch}{1.04}
  \begin{tabular}{clcccccccccc}
    \toprule
    \textbf{Model} & \textbf{Method}
    & \multicolumn{7}{c}{\textbf{ALFWorld}}
    & \textbf{WebShop}
    & \multicolumn{2}{c}{\textbf{ScienceWorld}} \\
    \cmidrule(lr){3-9}\cmidrule(lr){10-10}\cmidrule(lr){11-12}
    & & \textbf{Pick} & \textbf{Look} & \textbf{Clean} & \textbf{Heat} & \textbf{Cool} & \textbf{Pick2} & \cellcolor{alfcol}\textbf{All}
    & \cellcolor{webcol}\textbf{Total} & \cellcolor{scicol}\textbf{Seen} & \cellcolor{scicol}\textbf{Unseen} \\
    \midrule
    \multirow{7}{*}{Qwen2.5-7B} & CoT & 17.14 & 15.38 & 18.52 & 18.75 & 16.00 & 0.00 & \cellcolor{alfcol}14.29 & \cellcolor{webcol}5.10 & \cellcolor{scicol}3.09 & \cellcolor{scicol}4.63 \\
    & ReAct & 20.00 & 23.08 & 22.22 & 18.75 & 20.00 & 0.00 & \cellcolor{alfcol}17.14 & \cellcolor{webcol}15.28 & \cellcolor{scicol}8.24 & \cellcolor{scicol}9.93 \\
    & RAP & 40.00 & 15.38 & 33.33 & 6.25 & 32.00 & 20.83 & \cellcolor{alfcol}27.86 & \cellcolor{webcol}11.28 & \cellcolor{scicol}10.30 & \cellcolor{scicol}16.55 \\
    & ITP & 65.71 & 30.77 & 25.93 & 25.00 & 24.00 & \textbf{25.00} & \cellcolor{alfcol}35.71 & \cellcolor{webcol}20.10 & \cellcolor{scicol}\underline{16.49} & \cellcolor{scicol}\underline{17.88} \\
    & WM & \textbf{75.00} & 72.22 & \underline{54.84} & \underline{43.48} & 42.86 & 5.88 & \cellcolor{alfcol}49.05 & \cellcolor{webcol}41.57 & \cellcolor{scicol}11.86 & \cellcolor{scicol}12.32 \\
    & \method{} & \underline{70.83} & \textbf{88.89} & 45.16 & \textbf{47.83} & \textbf{71.43} & 17.65 & \cellcolor{alfcol}\underline{56.96} & \cellcolor{webcol}\textbf{46.90} & \cellcolor{scicol}15.46 & \cellcolor{scicol}16.58 \\
    & \method{}+Skill & 62.50 & \underline{83.30} & \textbf{64.52} & \textbf{47.83} & \underline{61.90} & \underline{23.53} & \cellcolor{alfcol}\textbf{57.27} & \cellcolor{webcol}\underline{46.44} & \cellcolor{scicol}\textbf{19.59} & \cellcolor{scicol}\textbf{20.38} \\
    \midrule
    \multirow{7}{*}{Qwen3-8B} & CoT & 14.29 & 15.38 & 14.81 & 12.50 & 12.00 & 12.50 & \cellcolor{alfcol}13.57 & \cellcolor{webcol}6.32 & \cellcolor{scicol}2.44 & \cellcolor{scicol}1.99 \\
    & ReAct & 25.71 & 7.69 & 22.22 & 12.50 & 12.00 & 25.00 & \cellcolor{alfcol}19.29 & \cellcolor{webcol}18.62 & \cellcolor{scicol}9.79 & \cellcolor{scicol}8.61 \\
    & RAP & 42.86 & 15.38 & 37.04 & 37.50 & 16.00 & 4.17 & \cellcolor{alfcol}28.57 & \cellcolor{webcol}12.40 & \cellcolor{scicol}15.46 & \cellcolor{scicol}\textbf{27.14} \\
    & ITP & \underline{82.86} & 23.08 & 25.93 & 12.50 & 16.00 & 54.17 & \cellcolor{alfcol}41.43 & \cellcolor{webcol}25.25 & \cellcolor{scicol}20.61 & \cellcolor{scicol}19.86 \\
    & WM & 70.83 & 38.89 & 41.94 & \underline{47.83} & 38.10 & \textbf{70.59} & \cellcolor{alfcol}51.36 & \cellcolor{webcol}50.00 & \cellcolor{scicol}20.10 & \cellcolor{scicol}20.63 \\
    & \method{} & 75.00 & \underline{50.00} & \underline{42.94} & \textbf{65.22} & \underline{42.86} & 58.82 & \cellcolor{alfcol}\underline{55.64} & \cellcolor{webcol}\underline{51.85} & \cellcolor{scicol} \underline{23.37} & 
    \cellcolor{scicol}19.91 \\
    & \method{}+Skill & \textbf{87.50} & \textbf{55.56} & \textbf{61.29} & \textbf{65.22} & \textbf{57.14} & \underline{64.17} & \cellcolor{alfcol}\textbf{65.24} & \cellcolor{webcol}\textbf{53.41} & \cellcolor{scicol}\textbf{24.23} & \cellcolor{scicol}\underline{21.85} \\
    \bottomrule
  \end{tabular}
  \endgroup
  \caption{
  \textbf{Main task-performance results across ALFWorld, WebShop, and ScienceWorld.}
  CoT, ReAct, RAP, and ITP are prompting or planning baselines; WM denotes the SFT-trained world-model agent; \method{} uses memory-augmented world modeling; and \method{}+Skill additionally uses policy-side world skill. Within each backbone block, best scores in each column are bolded, and second-best scores are underlined.
  }
  \label{tab:main-results}
\end{table*}

\subsection{Main Results}

\textbf{Takeaway \circnum{1}: Memory-augmented training consistently improves structured next-state fidelity.}
Table~\ref{tab:ssf-results} reports SSF for world-model predictions.
Mem-Aug RL obtains the best score in all nine model--benchmark cells, indicating that the gains are not tied to a single backbone or environment.
Compared with RL without memory, the largest improvements appear on WebShop, where preserving product identifiers, prices, and page fields is especially important: +0.175 for Llama3.2-1B, +0.070 for Qwen3-4B, and +0.135 for Qwen2.5-7B.
The improvements on ALFWorld and ScienceWorld are smaller but consistent, showing that \emph{world memory} also helps preserve state facts and transition rules.

Figure~\ref{fig:memory-dropout} provides a direct inference-time check.
When retrieved world memory blocks are removed from the prediction prompt, SSF decreases across ALFWorld, WebShop, and ScienceWorld.
This supports the interpretation that retrieved memory is used by the world model during imagination, rather than merely adding inert context.

\textbf{Takeaway \circnum{2}: Memory improves agent performance without policy training.}
Table~\ref{tab:main-results} reports task success.
Across both backbones, \method{} improves over the SFT-trained WM agent on ALFWorld overall and WebShop total, and adding world skill yields the strongest ALFWorld results.
On ScienceWorld, memory-augmented variants also improve over the corresponding SFT-trained WM agent in the Qwen2.5-7B block and on the Qwen3-8B seen split.
Because the policy model is frozen in all our agent variants, these gains come from memory-conditioned imagination and retrieved policy-side guidance rather than additional policy training.

\subsection{Sensitivity Analysis}
\begin{figure}[!t]
  \centering
  \includegraphics[width=\columnwidth]{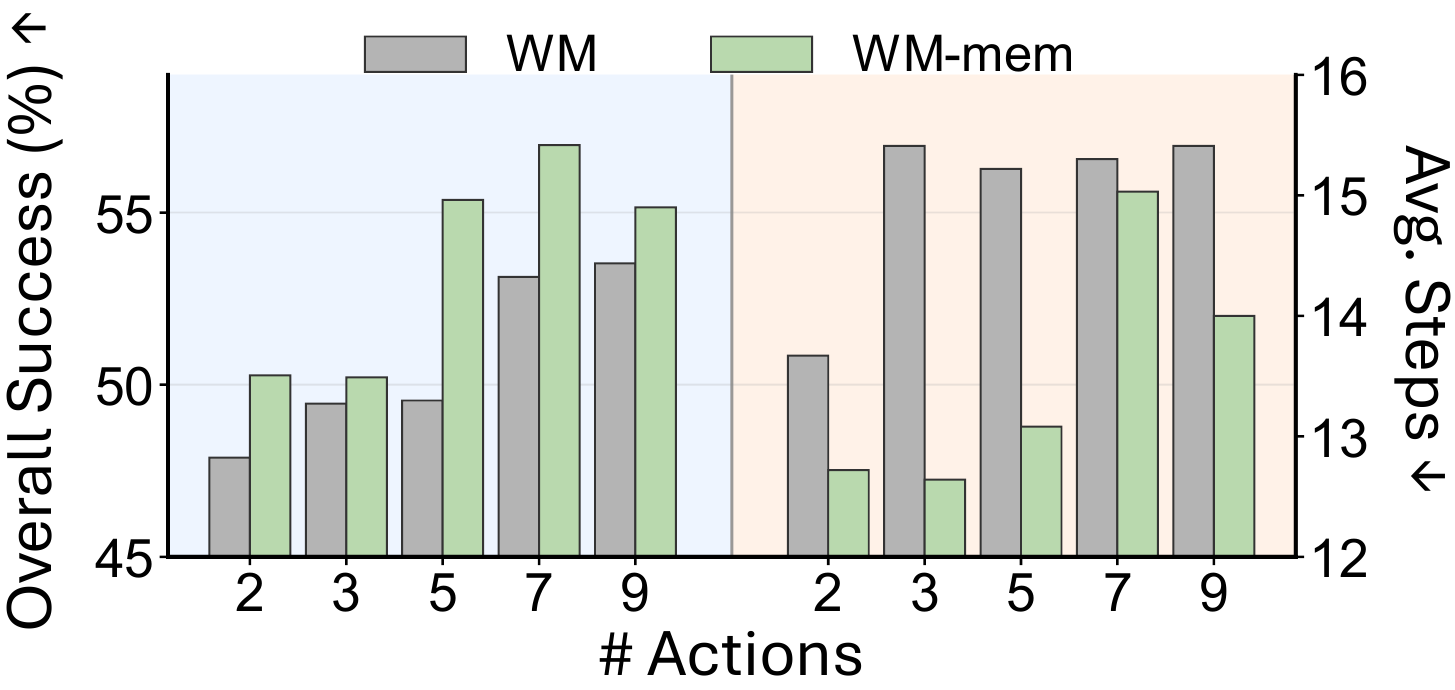}
  \caption{
  \textbf{Action-budget sensitivity on ALFWorld.}
  Memory-augmented world modeling improves success across candidate-action budgets while keeping successful trajectories comparable in length.
  }
  \label{fig:action-budget}
\end{figure}

\begin{figure}[!t]
  \centering
  \includegraphics[width=\columnwidth]{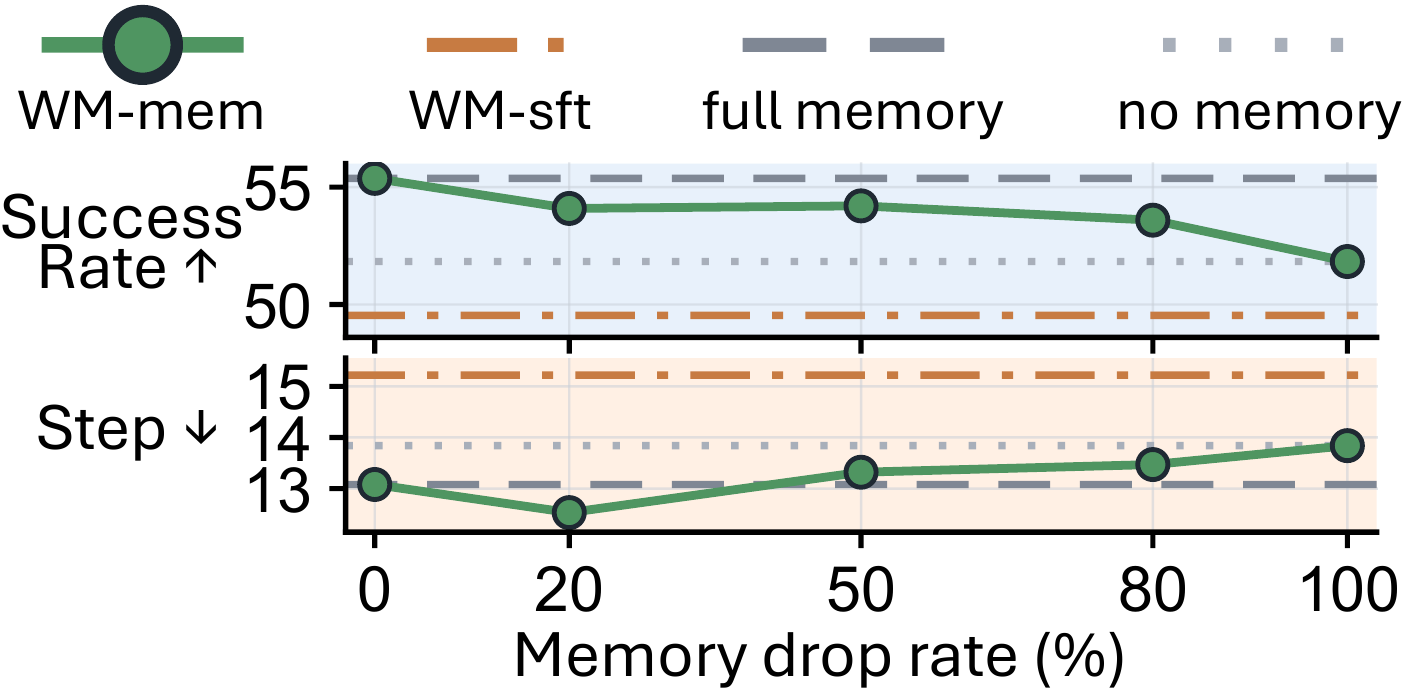}
  \caption{
  \textbf{Task-level memory-dropout sensitivity on ALFWorld.}
  With Qwen2.5-7B, full world memory gives the highest success rate, while heavier memory dropout lowers performance.
  }
  \label{fig:dropout-task}
\end{figure}

\textbf{Takeaway \circnum{3}: The gains are not an artifact of a single search budget or full-memory setting.}
Figure~\ref{fig:action-budget} varies the number of candidate actions considered by the planner.
Memory-augmented world modeling improves success across the tested budgets, including the default five-action setting.
At the same time, successful-task step counts are comparable to or lower than the SFT-trained WM setting.
Thus, the improvement is not simply obtained by exploring more actions; retrieved memory also helps the agent choose useful actions more efficiently.

Figure~\ref{fig:dropout-task} studies task-level memory removal.
Full \emph{world memory} gives the strongest ALFWorld success rate.
Moderate dropout retains most of the success rate and can reduce the number of steps for successful tasks, while heavier dropout lowers performance.
Since the all-drop setting still uses the RL-trained world model, it is not identical to the SFT reference; the gap between full memory and all-drop therefore isolates the contribution of retrieval at inference time.

\section{Conclusion}

We introduce \method{}, a memory-augmented text-based world model that uses \emph{world memory} to retrieve curated transition rules, state caches, and hard-to-predict facts during next-state imagination. To evaluate whether imagined states preserve behavior-critical information, we propose Structured State Fidelity (SSF), which compares predicted and gold states through benchmark-specific structured facts and fields. Across ALFWorld, WebShop, and ScienceWorld, memory-augmented training improves SSF over SFT and RL without memory. In downstream planning, combining \emph{world memory} with policy-side \emph{world skill} improves task success on ALFWorld and WebShop, while sensitivity analyses show that retrieved memory contributes useful information at inference time. Overall, the results show that externalized world memory is a practical mechanism for improving both state fidelity and downstream planning in text-based agents.

\bibliography{references}

@inproceedings{shridhar2021alfworld,
  title = {{ALFWorld}: Aligning Text and Embodied Environments for Interactive Learning},
  author = {Shridhar, Mohit and Yuan, Xingdi and C{\^o}t{\'e}, Marc-Alexandre and Bisk, Yonatan and Trischler, Adam and Hausknecht, Matthew},
  booktitle = {International Conference on Learning Representations},
  year = {2021},
  url = {https://arxiv.org/abs/2010.03768}
}

@inproceedings{wang2022scienceworld,
  title = {{ScienceWorld}: Is your Agent Smarter than a 5th Grader?},
  author = {Wang, Ruoyao and Jansen, Peter and C{\^o}t{\'e}, Marc-Alexandre and Ammanabrolu, Prithviraj},
  booktitle = {Proceedings of the 2022 Conference on Empirical Methods in Natural Language Processing},
  pages = {11279--11298},
  year = {2022},
  doi = {10.18653/v1/2022.emnlp-main.775},
  url = {https://aclanthology.org/2022.emnlp-main.775}
}

@inproceedings{yao2022webshop,
  title = {{WebShop}: Towards Scalable Real-World Web Interaction with Grounded Language Agents},
  author = {Yao, Shunyu and Chen, Howard and Yang, John and Narasimhan, Karthik},
  booktitle = {Advances in Neural Information Processing Systems},
  volume = {35},
  year = {2022},
  url = {https://arxiv.org/abs/2207.01206}
}

@article{ha2018worldmodels,
  title = {World Models},
  author = {Ha, David and Schmidhuber, J{\"u}rgen},
  journal = {arXiv preprint arXiv:1803.10122},
  year = {2018},
  url = {https://arxiv.org/abs/1803.10122}
}

@inproceedings{hafner2019dream,
  title = {Dream to Control: Learning Behaviors by Latent Imagination},
  author = {Hafner, Danijar and Lillicrap, Timothy and Ba, Jimmy and Norouzi, Mohammad},
  booktitle = {International Conference on Learning Representations},
  year = {2020},
  url = {https://openreview.net/forum?id=S1lOTC4tDS}
}

@article{qiao2024wkm,
  title={Agent Planning with World Knowledge Model},
  author={Shuofei Qiao and Runnan Fang and Ningyu Zhang and Yuqi Zhu and Xiang Chen and Shumin Deng and Yong Jiang and Pengjun Xie and Fei Huang and Huajun Chen},
  journal={ArXiv},
  year={2024},
  volume={abs/2405.14205},
  url = {https://arxiv.org/abs/2405.14205}
}

@article{li2025wordtoworld,
  title={From Word to World: Can Large Language Models be Implicit Text-based World Models?},
  author={Yixia Li and Hongru Wang and Jiahao Qiu and Zhenfei Yin and Dongdong Zhang and Cheng Qian and Zeping Li and Po-shan Cathy Ma and Guanhua Chen and Heng Ji and Mengdi Wang},
  journal={ArXiv},
  year={2025},
  volume={abs/2512.18832},
  url = {https://arxiv.org/abs/2512.18832}
}

@article{liu2026itp,
  title = {Imagine-then-Plan: Agent Learning from Adaptive Lookahead with World Models},
  author = {Liu, Youwei and Wang, Jian and Wang, Hanlin and Guo, Beichen and Li, Wenjie},
  journal = {arXiv preprint arXiv:2601.08955},
  year = {2026},
  volume = {abs/2601.08955},
  url = {https://arxiv.org/abs/2601.08955}
}

@article{huang2026behaviorconsistency,
  title = {Beyond State Consistency: Behavior Consistency in Text-Based World Models},
  author = {Huang, Youling and Chen, Guanqiao and Yao, Junchi and Wang, Lu and Yang, Fangkai and Du, Chao and Zhao, ChenZhuo and Zhao, Pu and Lin, Qingwei and Rajmohan, Saravan and Zhang, Dongmei},
  journal = {arXiv preprint arXiv:2604.13824},
  year = {2026},
  url = {https://arxiv.org/abs/2604.13824}
}

@inproceedings{lewis2020rag,
 author = {Lewis, Patrick and Perez, Ethan and Piktus, Aleksandra and Petroni, Fabio and Karpukhin, Vladimir and Goyal, Naman and K\"{u}ttler, Heinrich and Lewis, Mike and Yih, Wen-tau and Rockt\"{a}schel, Tim and Riedel, Sebastian and Kiela, Douwe},
 booktitle = {Advances in Neural Information Processing Systems},
 editor = {H. Larochelle and M. Ranzato and R. Hadsell and M.F. Balcan and H. Lin},
 pages = {9459--9474},
 publisher = {Curran Associates, Inc.},
 title = {Retrieval-Augmented Generation for Knowledge-Intensive NLP Tasks},
 url = {https://proceedings.neurips.cc/paper_files/paper/2020/file/6b493230205f780e1bc26945df7481e5-Paper.pdf},
 volume = {33},
 year = {2020}
}

@inproceedings{borgeaud2021retro,
  author    = {Borgeaud, Sebastian and Mensch, Arthur and Hoffmann, Jordan and Cai, Trevor and Rutherford, Eliza and Millican, Katie and Van Den Driessche, George Bm and Lespiau, Jean-Baptiste and Damoc, Bogdan and Clark, Aidan and De Las Casas, Diego and Guy, Aurelia and Menick, Jacob and Ring, Roman and Hennigan, Tom and Huang, Saffron and Maggiore, Loren and Jones, Chris and Cassirer, Albin and Brock, Andy and Paganini, Michela and Irving, Geoffrey and Vinyals, Oriol and Osindero, Simon and Simonyan, Karen and Rae, Jack and Elsen, Erich and Sifre, Laurent},
  title     = {Improving Language Models by Retrieving from Trillions of Tokens},
  booktitle = {Proceedings of the 39th International Conference on Machine Learning},
  series    = {Proceedings of Machine Learning Research},
  volume    = {162},
  pages     = {2206--2240},
  year      = {2022},
  publisher = {PMLR},
  url       = {https://proceedings.mlr.press/v162/borgeaud22a.html}
}

@article{wang2023longmem,
  title = {Augmenting Language Models with Long-Term Memory},
  author = {Wang, Weizhi and Dong, Li and Cheng, Hao and Liu, Xiaodong and Yan, Xifeng and Gao, Jianfeng and Wei, Furu},
  journal = {arXiv preprint arXiv:2306.07174},
  year = {2023},
  url = {https://arxiv.org/abs/2306.07174}
}

@article{zhong2023memorybank,
  title = {{MemoryBank}: Enhancing Large Language Models with Long-Term Memory},
  author = {Zhong, Wanjun and Guo, Lianghong and Gao, Qiqi and Ye, He and Wang, Yanlin},
  journal = {arXiv preprint arXiv:2305.10250},
  year = {2023},
  url = {https://arxiv.org/abs/2305.10250}
}

@article{xia2026skillrl,
  title = {{SkillRL}: Evolving Agents via Recursive Skill-Augmented Reinforcement Learning},
  author = {Xia, Peng and Chen, Jianwen and Wang, Hanyang and Liu, Jiaqi and Zeng, Kaide and Wang, Yu and Han, Siwei and Zhou, Yiyang and Zhao, Xujiang and Chen, Haifeng and Zheng, Zeyu and Xie, Cihang and Yao, Huaxiu},
  journal = {arXiv preprint arXiv:2602.08234},
  year = {2026},
  doi = {10.48550/arXiv.2602.08234},
  url = {https://arxiv.org/abs/2602.08234}
}

@inproceedings{wei2022cot,
  title = {Chain-of-Thought Prompting Elicits Reasoning in Large Language Models},
  author = {Wei, Jason and Wang, Xuezhi and Schuurmans, Dale and Bosma, Maarten and Ichter, Brian and Xia, Fei and Chi, Ed H. and Le, Quoc V. and Zhou, Denny},
  booktitle = {Advances in Neural Information Processing Systems},
  volume = {35},
  pages = {24824--24837},
  year = {2022},
  url = {https://arxiv.org/abs/2201.11903}
}

@inproceedings{yao2023react,
  title = {{ReAct}: Synergizing Reasoning and Acting in Language Models},
  author = {Yao, Shunyu and Zhao, Jeffrey and Yu, Dian and Du, Nan and Shafran, Izhak and Narasimhan, Karthik and Cao, Yuan},
  booktitle = {International Conference on Learning Representations},
  year = {2023},
  url = {https://openreview.net/forum?id=WE_vluYUL-X}
}

@inproceedings{hao2023rap,
  title = {Reasoning with Language Model is Planning with World Model},
  author = {Hao, Shibo and Gu, Yi and Ma, Haodi and Hong, Joshua and Wang, Zhen and Wang, Daisy Zhe and Hu, Zhiting},
  booktitle = {Proceedings of the 2023 Conference on Empirical Methods in Natural Language Processing},
  pages = {8154--8173},
  year = {2023},
  doi = {10.18653/v1/2023.emnlp-main.507},
  url = {https://aclanthology.org/2023.emnlp-main.507}
}

@article{sutton1991dyna,
  author  = {Richard S. Sutton},
  title   = {Dyna, an Integrated Architecture for Learning, Planning, and Reacting},
  journal = {ACM SIGART Bulletin},
  volume  = {2},
  number  = {4},
  pages   = {160--163},
  year    = {1991},
  doi     = {10.1145/122344.122377}
}

@inproceedings{janner2019mbpo,
  title = {When to Trust Your Model: Model-Based Policy Optimization},
  author = {Janner, Michael and Fu, Justin and Zhang, Marvin and Levine, Sergey},
  booktitle = {Advances in Neural Information Processing Systems},
  volume = {32},
  year = {2019},
  url = {https://proceedings.neurips.cc/paper/2019/hash/5faf461eff3099671ad63c6f3f094f7f-Abstract.html}
}

@article{schrittwieser2020muzero,
  title = {Mastering {Atari}, {Go}, Chess and Shogi by Planning with a Learned Model},
  author = {Schrittwieser, Julian and Antonoglou, Ioannis and Hubert, Thomas and Simonyan, Karen and Sifre, Laurent and Schmitt, Simon and Guez, Arthur and Lockhart, Edward and Hassabis, Demis and Graepel, Thore and Lillicrap, Timothy and Silver, David},
  journal = {Nature},
  volume = {588},
  pages = {604--609},
  year = {2020},
  doi = {10.1038/s41586-020-03051-4},
  url = {https://www.nature.com/articles/s41586-020-03051-4}
}

@inproceedings{cote2018textworld,
  title={TextWorld: A Learning Environment for Text-based Games},
  author={C{\^o}t{\'e}, Marc-Alexandre and K{\'a}d{\'a}r, {\'A}kos and Yuan, Xingdi and Kybartas, Ben and Barnes, Tavian and Fine, Emery and Moore, James and Tao, Ruo Yu and Hausknecht, Matthew and El Asri, Layla and Adada, Mahmoud and Tay, Wendy and Trischler, Adam},
  booktitle={CGW@IJCAI},
  year={2018},
  url={https://arxiv.org/abs/1806.11532}
}

@article{hausknecht2020jericho,
  author  = {Matthew Hausknecht and Prithviraj Ammanabrolu
             and Marc-Alexandre C{\^o}t{\'e} and Xingdi Yuan},
  title   = {Interactive Fiction Games: A Colossal Adventure},
  journal = {Proceedings of the AAAI Conference on Artificial Intelligence},
  volume  = {34},
  number  = {5},
  pages   = {7903--7910},
  year    = {2020},
  doi     = {10.1609/aaai.v34i05.6297}
}

@inproceedings{narasimhan2015textgames,
  title = {Language Understanding for Text-Based Games Using Deep Reinforcement Learning},
  author = {Narasimhan, Karthik and Kulkarni, Tejas and Barzilay, Regina},
  booktitle = {Proceedings of the 2015 Conference on Empirical Methods in Natural Language Processing},
  pages = {1--11},
  year = {2015},
  doi = {10.18653/v1/D15-1001},
  url = {https://aclanthology.org/D15-1001/}
}

@article{ahn2022saycan,
  title = {Do As I Can, Not As I Say: Grounding Language in Robotic Affordances},
  author = {Ahn, Michael and Brohan, Anthony and Brown, Noah and Chebotar, Yevgen and Cortes, Omar and David, Byron and Finn, Chelsea and Fu, Chuyuan and Gopalakrishnan, Keerthana and Hausman, Karol and Herzog, Alex and Ho, Daniel and Hsu, Jasmine and Ibarz, Julian and Ichter, Brian and Irpan, Alex and Jang, Eric and Jauregui Ruano, Rosario and Jeffrey, Kyle and Jesmonth, Sally and Joshi, Nikhil J and Julian, Ryan and Kalashnikov, Dmitry and Kuang, Yuheng and Lee, Kuang-Huei and Levine, Sergey and Lu, Yao and Luu, Linda and Parada, Carolina and Pastor, Peter and Quiambao, Jornell and Rao, Kanishka and Rettinghouse, Jarek and Reyes, Diego and Sermanet, Pierre and Sievers, Nicolas and Tan, Clayton and Toshev, Alexander and Vanhoucke, Vincent and Xia, Fei and Xiao, Ted and Xu, Peng and Xu, Sichun and Yan, Mengyuan and Zeng, Andy},
  journal = {arXiv preprint arXiv:2204.01691},
  year = {2022},
  doi = {10.48550/arXiv.2204.01691},
  url = {https://arxiv.org/abs/2204.01691}
}

@article{wang2023voyager,
  title = {{Voyager}: An Open-Ended Embodied Agent with Large Language Models},
  author = {Wang, Guanzhi and Xie, Yuqi and Jiang, Yunfan and Mandlekar, Ajay and Xiao, Chaowei and Zhu, Yuke and Fan, Linxi and Anandkumar, Anima},
  journal = {arXiv preprint arXiv:2305.16291},
  year = {2023},
  doi = {10.48550/arXiv.2305.16291},
  url = {https://arxiv.org/abs/2305.16291}
}

@inproceedings{park2023generativeagents,
  title = {Generative Agents: Interactive Simulacra of Human Behavior},
  author = {Park, Joon Sung and O'Brien, Joseph C. and Cai, Carrie J. and Morris, Meredith Ringel and Liang, Percy and Bernstein, Michael S.},
  booktitle = {Proceedings of the 36th Annual ACM Symposium on User Interface Software and Technology},
  year = {2023},
  doi = {10.1145/3586183.3606763},
  url = {https://doi.org/10.1145/3586183.3606763}
}

@inproceedings{shinn2023reflexion,
 author = {Shinn, Noah and Cassano, Federico and Gopinath, Ashwin and Narasimhan, Karthik and Yao, Shunyu},
 booktitle = {Advances in Neural Information Processing Systems},
 editor = {A. Oh and T. Naumann and A. Globerson and K. Saenko and M. Hardt and S. Levine},
 pages = {8634--8652},
 publisher = {Curran Associates, Inc.},
 title = {Reflexion: language agents with verbal reinforcement learning},
 url = {https://proceedings.neurips.cc/paper_files/paper/2023/file/1b44b878bb782e6954cd888628510e90-Paper-Conference.pdf},
 volume = {36},
 year = {2023}
}

@inproceedings{papineni2002bleu,
  title = {{BLEU}: A Method for Automatic Evaluation of Machine Translation},
  author = {Papineni, Kishore and Roukos, Salim and Ward, Todd and Zhu, Wei-Jing},
  booktitle = {Proceedings of the 40th Annual Meeting of the Association for Computational Linguistics},
  pages = {311--318},
  year = {2002},
  doi = {10.3115/1073083.1073135},
  url = {https://aclanthology.org/P02-1040/}
}

@inproceedings{lin2004rouge,
  title = {{ROUGE}: A Package for Automatic Evaluation of Summaries},
  author = {Lin, Chin-Yew},
  booktitle = {Text Summarization Branches Out},
  pages = {74--81},
  year = {2004},
  url = {https://aclanthology.org/W04-1013/}
}

@inproceedings{zhang2020bertscore,
  title = {{BERTScore}: Evaluating Text Generation with {BERT}},
  author = {Zhang, Tianyi and Kishore, Varsha and Wu, Felix and Weinberger, Kilian Q. and Artzi, Yoav},
  booktitle = {International Conference on Learning Representations},
  year = {2020},
  url = {https://arxiv.org/abs/1904.09675}
}

@inproceedings{min2023factscore,
    title = "{FA}ct{S}core: Fine-grained Atomic Evaluation of Factual Precision in Long Form Text Generation",
    author = "Min, Sewon  and
      Krishna, Kalpesh  and
      Lyu, Xinxi  and
      Lewis, Mike  and
      Yih, Wen-tau  and
      Koh, Pang  and
      Iyyer, Mohit  and
      Zettlemoyer, Luke  and
      Hajishirzi, Hannaneh",
    editor = "Bouamor, Houda  and
      Pino, Juan  and
      Bali, Kalika",
    booktitle = "Proceedings of the 2023 Conference on Empirical Methods in Natural Language Processing",
    month = dec,
    year = "2023",
    address = "Singapore",
    publisher = "Association for Computational Linguistics",
    url = "https://aclanthology.org/2023.emnlp-main.741/",
    doi = "10.18653/v1/2023.emnlp-main.741",
    pages = "12076--12100"
}

@inproceedings{karpukhin2020dpr,
  title = {Dense Passage Retrieval for Open-Domain Question Answering},
  author = {Karpukhin, Vladimir and Oguz, Barlas and Min, Sewon and Lewis, Patrick and Wu, Ledell and Edunov, Sergey and Chen, Danqi and Yih, Wen-tau},
  booktitle = {Proceedings of the 2020 Conference on Empirical Methods in Natural Language Processing},
  pages = {6769--6781},
  year = {2020},
  doi = {10.18653/v1/2020.emnlp-main.550},
  url = {https://aclanthology.org/2020.emnlp-main.550/}
}

@inproceedings{guu2020realm,
  title = {{REALM}: Retrieval-Augmented Language Model Pre-Training},
  author = {Guu, Kelvin and Lee, Kenton and Tung, Zora and Pasupat, Panupong and Chang, Ming-Wei},
  booktitle = {Proceedings of the 37th International Conference on Machine Learning},
  year = {2020},
  url = {https://proceedings.mlr.press/v119/guu20a.html}
}

@inproceedings{izacard2021fid,
  title = {Leveraging Passage Retrieval with Generative Models for Open Domain Question Answering},
  author = {Izacard, Gautier and Grave, Edouard},
  booktitle = {Proceedings of the 16th Conference of the European Chapter of the Association for Computational Linguistics},
  pages = {874--880},
  year = {2021},
  doi = {10.18653/v1/2021.eacl-main.74},
  url = {https://aclanthology.org/2021.eacl-main.74/}
}

@inproceedings{
bi2026echorl,
title={Echo{RL}: Reinforcement Learning via Rollout Echoing},
author = {Jinhe Bi and Aniri and Minglai Yang and Xingcheng Zhou and Wenke Huang and Sikuan Yan and Yujun Wang and Zixuan Cao and Michael Färber and Xun Xiao and Volker Tresp and Yunpu Ma}
,
booktitle={Forty-third International Conference on Machine Learning},
year={2026},
url={https://openreview.net/forum?id=A6az59SGtF}
}

@inproceedings{
bi2026the,
title={The Geometry of Reasoning: Self-Evaluation via Layerwise Trajectory Evolution},
author = {Jinhe Bi and Danqi Yan and Yifan Wang and Wenke Huang and Haokun Chen and Guancheng Wan and Mang Ye and Xun Xiao and Hinrich Schuetze and Volker Tresp and Yunpu Ma},
booktitle={Forty-third International Conference on Machine Learning},
year={2026},
url={https://openreview.net/forum?id=WQyrwQwzmK}
}

@misc{huang2025loongsynthesizelongchainofthoughts,
      title={Loong: Synthesize Long Chain-of-Thoughts at Scale through Verifiers}, 
      author={Xingyue Huang and Rishabh and Gregor Franke and Ziyi Yang and Jiamu Bai and Weijie Bai and Jinhe Bi and Zifeng Ding and Yiqun Duan and Chengyu Fan and Wendong Fan and Xin Gao and Ruohao Guo and Yuan He and Zhuangzhuang He and Xianglong Hu and Neil Johnson and Bowen Li and Fangru Lin and Siyu Lin and Tong Liu and Yunpu Ma and Hao Shen and Hao Sun and Beibei Wang and Fangyijie Wang and Hao Wang and Haoran Wang and Yang Wang and Yifeng Wang and Zhaowei Wang and Ziyang Wang and Yifan Wu and Zikai Xiao and Chengxing Xie and Fan Yang and Junxiao Yang and Qianshuo Ye and Ziyu Ye and Guangtao Zeng and Yuwen Ebony Zhang and Zeyu Zhang and Zihao Zhu and Bernard Ghanem and Philip Torr and Guohao Li},
      year={2025},
      eprint={2509.03059},
      archivePrefix={arXiv},
      primaryClass={cs.LG},
      url={https://arxiv.org/abs/2509.03059}, 
}

@misc{zhao2026nl2codestructuredsurveymultimodal,
      title={Beyond NL2Code: A Structured Survey of Multimodal Code Intelligence}, 
      author={Xuanle Zhao and Qiushi Sun and Jingyu Xiao and Xuexin Liu and Haoyue Yang and Qiaosheng Chen and Xianzhen Luo and Jing Huang and Yufeng Zhong and Lei Chen and Shuai Fu and Zhenlin Wei and Jinhe Bi and Lei Jiang and Haibo Qiu and Siqi Yang and Peng Shi and Jian Hu and Zhixiong Zeng},
      year={2026},
      eprint={2606.15932},
      archivePrefix={arXiv},
      primaryClass={cs.CL},
      url={https://arxiv.org/abs/2606.15932}, 
}

@misc{mid,
      title={Reinforcement Mid-Training}, 
      author={Yijun Tian and Shaoyu Chen and Zhichao Xu and Yawei Wang and Jinhe Bi and Peng Han and Wei Wang},
      year={2025},
      eprint={2509.24375},
      archivePrefix={arXiv},
      primaryClass={cs.CL},
      url={https://arxiv.org/abs/2509.24375}, 
}

@inproceedings{jansen2023textworldexpress,
  title = {{TextWorldExpress}: Simulating Text Games at One Million Steps Per Second},
  author = {Jansen, Peter and Cote, Marc-alexandre},
  booktitle = {Proceedings of the 17th Conference of the European Chapter of the Association for Computational Linguistics: System Demonstrations},
  pages = {169--177},
  year = {2023},
  publisher = {Association for Computational Linguistics},
  doi = {10.18653/v1/2023.eacl-demo.20},
  url = {https://aclanthology.org/2023.eacl-demo.20/}
}

@article{wang2026ascd,
  title = {{ASCD}: Attention-Steerable Contrastive Decoding for Reducing Hallucination in {MLLM}},
  author = {Wang, Yujun and Aniri and Bi, Jinhe and Pirk, Soren and Ma, Yunpu},
  journal = {Proceedings of the AAAI Conference on Artificial Intelligence},
  volume = {40},
  number = {12},
  pages = {10306--10314},
  year = {2026},
  doi = {10.1609/aaai.v40i12.38000},
  url = {https://ojs.aaai.org/index.php/AAAI/article/view/38000}
}

\clearpage
\appendix
\section{Benchmark and Evaluation Setup}
\label{sec:appendix-benchmark-setup}

\paragraph{Common protocol.}
All experiments use the text interface provided by each benchmark. At each decision step, the agent receives a textual observation and issues a textual action from the benchmark action interface. For downstream task evaluation, we use the task success or task score returned by the original benchmark environment. Within each benchmark, all compared methods use the same task instances, observation format, action interface, evaluation budget, decoding settings, and planner interface.

For world-model evaluation, each example is an environment transition. Given the current textual state $s_t$, optional interaction history $h_t$, and candidate action $a_t$, the world model predicts an imagined next state $\hat{s}_{t+1}$. The benchmark environment provides the gold next state $s_{t+1}$ after executing $a_t$, which is used only for evaluation. We report Structured State Fidelity (SSF) as the primary next-state metric; exact match and word F1 are used only in the metric stress test to illustrate why surface metrics are insufficient.

\begin{table}[t]
  \centering
  \small
  \setlength{\tabcolsep}{7pt}
  \begin{tabular}{ccc}
    \toprule
    \textbf{Domain} & \textbf{Training} & \textbf{Test} \\
    \midrule
    ALFWorld & 3,119 & 139 \\
    ScienceWorld & 1,483 & Seen: 194; Unseen: 211 \\
    WebShop & 1,824 & 200 \\
    \bottomrule
  \end{tabular}
  \caption{Dataset split statistics for the three evaluation domains.}
  \label{tab:dataset-splits}
\end{table}

\paragraph{Prompting baselines in Table~\ref{tab:main-results}.}
The first four rows in each backbone block of Table~\ref{tab:main-results} reproduce the prompting-based baseline results from the Imagine-then-Plan study \citep{liu2026itp}. \textbf{CoT} prompts the agent to produce step-by-step rationales before acting, following chain-of-thought prompting \citep{wei2022cot}. \textbf{ReAct} interleaves reasoning traces with environment actions, so the agent repeatedly reasons about the current observation and then issues an action \citep{yao2023react}. \textbf{RAP} treats the LLM as both a policy and a world model, and uses Monte Carlo Tree Search to plan over imagined action--state continuations \citep{hao2023rap}. \textbf{ITP} denotes the training-free ITPI variant from \citet{liu2026itp}: the agent selects an adaptive lookahead horizon, rolls out imagined futures with a learned world model, and conditions the next action on a reflection over those imagined futures. We use the label ITP for compactness in the table; it should not be confused with the reinforced ITPR variant reported in the source paper.

\paragraph{ALFWorld.}
ALFWorld \citep{shridhar2021alfworld} is a text-based embodied household benchmark aligned with ALFRED-style goals. An agent navigates rooms, inspects receptacles, manipulates objects, and completes goal-conditioned household tasks. Our evaluation reports the six task categories shown in the main results table: pick-and-place, look-at-object, clean-and-place, heat-and-place, cool-and-place, and pick-two-object tasks, together with the overall success rate. For state-prediction evaluation, the gold state is the textual observation returned by ALFWorld after executing the candidate action. This setting stresses whether a world model preserves object identities, receptacle relations, object states, and action preconditions.

\paragraph{ScienceWorld.}
ScienceWorld \citep{wang2022scienceworld} is an interactive text environment for elementary-school science tasks. Episodes require an agent to manipulate objects, containers, devices, and substances while satisfying procedural and scientific constraints. The benchmark distinguishes seen and unseen task settings, which we keep separate in the downstream table when reporting non-memory baselines. For world-model evaluation, each transition prediction is scored against the next textual observation produced by the simulator. This benchmark stresses preservation of entities, device states, container contents, numeric measurements, recipes, and scientific process feedback.

\paragraph{WebShop.}
WebShop \citep{yao2022webshop} is a simulated e-commerce website built from real-world product data and crowd-sourced shopping instructions. Given a user request, an agent searches, navigates result pages and product pages, selects options, and attempts to purchase a matching product. We report the benchmark's total task score in the main table. For state-prediction evaluation, the gold next state is the next webpage observation after an action such as searching, opening a product, selecting an option, navigating pages, or buying. This benchmark stresses preservation of page type, product identifiers, titles, prices, option groups, navigation affordances, and the user's stated constraints.

\section{Structured State Fidelity Details}
\label{sec:appendix-ssf}

This appendix gives additional implementation details for Structured State Fidelity (SSF), the proposed metric introduced in Section~\ref{sec:ssf}. SSF evaluates a predicted next state by first extracting a structured representation from both the prediction and the gold state, then computing similarity over the extracted state elements. This differs from exact match and word F1, which operate directly on surface strings.

\subsection{ALFWorld SSF}

For ALFWorld, SSF converts each observation into a set of canonical facts using a tiny LLM extractor. The extractor normalizes surface descriptions into object--receptacle relations (e.g., \texttt{ON} and \texttt{IN}), empty-receptacle facts (\texttt{EMPTY}), object-state facts (\texttt{STATE} with values such as open, closed, heated, cooled, cleaned, dirty, or sliced), and task-specific facts such as \texttt{CLEANED}. It preserves object and receptacle identities, including numeric indices, because these identities are behavior-critical in ALFWorld.

Given predicted facts $\hat{F}$ and gold facts $F$, the ALFWorld score is fact-set F1. We compute precision, recall, and SSF as
\begin{equation}
\begin{gathered}
P = \frac{|\hat{F} \cap F|}{|\hat{F}|},
\qquad
R = \frac{|\hat{F} \cap F|}{|F|}, \\
\mathrm{SSF}_{\textsc{ALFWorld}}
= \frac{2PR}{P + R}.
\end{gathered}
\end{equation}

\subsection{ScienceWorld SSF}

For ScienceWorld, SSF follows the same fact-set F1 design, but uses a hybrid extractor. A deterministic parser first handles frequent ScienceWorld observation patterns, including room and location descriptions, visible objects, surface and container relations, empty containers, door and device states, movement or focus feedback, electrical connections, thermometer readings, waiting and pouring feedback, recipe text, ambiguous action menus, and short fallback messages. The resulting fact schema groups these observations into location facts, visibility and containment facts, object-state facts, device and measurement facts, and task-feedback facts. If the local parser cannot extract facts, a small LLM extractor is used as a fallback.

We use the hybrid parser rather than a pure LLM extractor because metric extraction itself should not repair the prediction being evaluated. In pilot checks, a pure LLM-only extractor sometimes over-completed corrupted predictions into gold-like facts, producing a perturbation maximum of $1.000$ even when the prediction had been deliberately corrupted. The hybrid extractor anchors frequent ScienceWorld templates with deterministic rules and invokes the small LLM only as a fallback. With cache disabled on 64 stratified ScienceWorld samples, the hybrid metric gives identity comparisons a mean/min/max score of $1.000/1.000/1.000$, while simple perturbations receive a mean score of $0.008$ and a maximum of $0.250$. This supports the intended behavior: faithful states are preserved, but corrupted facts are not silently repaired by the extractor.

ScienceWorld also illustrates why surface metrics can hide meaningful improvements. On the test split, a Qwen3-4B world model using only the current observation obtained word F1 $0.962$ and ScienceWorld SSF $0.848$. With ten steps of history, word F1 increased modestly to $0.973$, while ScienceWorld SSF increased to $0.910$. This suggests that word overlap is already near saturation, whereas structured state fidelity remains sensitive to improvements in behavior-relevant facts.

\subsection{WebShop SSF}

For WebShop, observations are structured pages rather than generic fact lists. The metric first classifies each state as a search-result page, product-detail page, terminal or completion page, or unknown page. For search-result pages, it extracts page number, total result count, navigation buttons, top-$k$ product identifiers, product titles, and prices, with $k=3$ in our experiments. For product-detail pages, it extracts title, price, option keys, option groups, and page affordances such as description, features, reviews, and buy-now buttons.

The WebShop score is a weighted structured similarity. Categorical fields use exact match, product identifiers and option sets use set F1, titles use lexical F1, and prices use a relative-error score. The final page score is the weighted sum of field scores, clipped to $[0,1]$. For search-result pages, the weights are $0.100$ for page type, page number, and total result count; $0.050$ for each navigation affordance (previous, next, and back-to-search); $0.300$ for top-$k$ product identifiers; $0.150$ for product titles; and $0.100$ for prices. For product-detail pages, the weights are $0.150$ for page type, $0.200$ for title, $0.150$ for price, $0.100$ each for option keys and option groups, and $0.050$ each for previous, back-to-search, description, features, reviews, and buy-now affordances. This design rewards preserving product identity and actionable attributes rather than merely matching the surrounding page text.

\subsection{Metric Stress Test}
\label{sec:appendix-metric-stress-test}

We further validate the motivation for SSF with a controlled metric stress test. The test samples real gold next states from ALFWorld, ScienceWorld, and WebShop, then constructs synthetic predictions that isolate different metric behaviors. We compare three metrics: normalized exact match (EM), token-count word F1 over normalized text, and SSF. The stress test targets 50 examples per perturbation class and scores SSF with the benchmark-specific structured metrics described above.

The perturbations are designed to test two desiderata. First, a metric should be robust to meaning-preserving paraphrase or layout changes. Second, it should be sensitive to behavior-critical corruptions, including object or location swaps, entity swaps, identifier changes, price or numeric changes, option deletion, and state flips. Figure~\ref{fig:ssf-motivation} summarizes the aggregate pattern in the main text. EM fails the paraphrase condition because all paraphrases receive zero exact-match score. Word F1 remains high even under fact corruption: object/location/entity swaps average 0.925 word F1, and ID/price/numeric/option corruptions average 0.934 word F1. SSF separates the cases more sharply: paraphrase/layout changes retain score 1.000, while object/location/entity swaps drop to 0.151 and state flips drop to 0.393.

The same pattern holds within individual benchmarks. In ALFWorld, location swaps receive word F1 0.920 but SSF 0.000, showing that token overlap can miss an entirely wrong receptacle or room. In ScienceWorld, entity and numeric swaps similarly retain word F1 above 0.920 while receiving SSF 0.000. In WebShop, product identifier swaps receive word F1 0.989 but SSF 0.450 because the page text is almost unchanged while the top product identities are wrong. These results support the central motivation for SSF: world-model evaluation should measure preservation of structured state facts, not only lexical overlap.

\subsection{Exact Extraction Prompts and Parser Rules}
\label{sec:appendix-ssf-extraction-rules}

We next provide the exact extraction interfaces used by SSF. The extractors process the predicted and gold next states independently. ALFWorld uses a constrained canonical-fact prompt; ScienceWorld first applies deterministic rules and invokes a constrained fallback prompt only if no local rule produces a fact; WebShop uses a deterministic page parser throughout.

\subsubsection{ALFWorld Canonical-Fact Prompt}

The complete instruction passed to the ALFWorld extractor is shown below. The placeholder \texttt{\{observation\}} is replaced by one predicted or gold next-state observation.

\begin{promptbox}[\textbf{Exact ALFWorld SSF Extraction Prompt}]
You are an information extraction system for ALFWorld observations.

Your task is to convert ONE observation into a canonical fact list.

You must normalize semantically equivalent expressions into the SAME canonical fact format.

You must follow the canonicalization rules exactly.

## Canonical fact rules

1. Visibility on a surface:
- "On the X, you see a A, a B, and a C."
  -> ON(X, A)
  -> ON(X, B)
  -> ON(X, C)

2. Visibility inside a container:
- "In the X, you see a A, a B, and a C."
  -> IN(X, A)
  -> IN(X, B)
  -> IN(X, C)

3. Empty observations:
The following expressions must all be normalized to EXACTLY:
- EMPTY(X)

Examples:
- "The X is empty." -> EMPTY(X)
- "On the X, you see nothing." -> EMPTY(X)
- "In the X, you see nothing." -> EMPTY(X)
- "Inside, you see nothing." -> EMPTY(X), after resolving the pronoun to the correct entity

4. State observations:
- "The X is open." -> STATE(X, open)
- "The X is closed." -> STATE(X, closed)
- "The X is heated." -> STATE(X, heated)
- "The X is cooled." -> STATE(X, cooled)
- "The X is cleaned." -> STATE(X, cleaned)
- "The X is dirty." -> STATE(X, dirty)
- "The X is sliced." -> STATE(X, sliced)

5. Cleaning actions:
If the observation says an object is cleaned using a sinkbasin or similar receptacle, the ONLY valid canonical fact is:
- CLEANED(OBJECT, RECEPTACLE)

Examples:
- "You clean the dishsponge 2 using the sinkbasin 1."
  -> CLEANED(dishsponge 2, sinkbasin 1)
- "You clean the cloth 1 using the sinkbasin 1."
  -> CLEANED(cloth 1, sinkbasin 1)

Never output CLEAN(...).
Always output CLEANED(...).

6. Observations containing both an action sentence and a resulting world-state sentence:
Only keep the resulting world-state facts.
Do NOT output redundant action facts.

Examples:
- "You open the fridge 1. The fridge 1 is open."
  -> STATE(fridge 1, open)

- "You open the drawer 1. The drawer 1 is open. In it, you see a fork 2."
  -> STATE(drawer 1, open)
  -> IN(drawer 1, fork 2)

- "You open the fridge 1. The fridge 1 is open. Inside, you see nothing."
  -> STATE(fridge 1, open)
  -> EMPTY(fridge 1)

7. Pronoun resolution:
Resolve pronouns such as:
- "Inside"
- "In it"
- "On it"

to the explicit receptacle/entity mentioned earlier in the same observation.

8. Identity preservation:
Object identity must be preserved exactly, including indices.
Examples:
- candle 1 != candle 2
- spraybottle 1 != spraybottle 2

Receptacle identity must be preserved exactly, including indices.
Examples:
- drawer 1 != drawer 2
- fridge 1 != fridge 2

## Output format

Return the facts ONLY between the two markers below:

BEGIN_FACTS
<one complete canonical fact per line>
END_FACTS

Strict output rules:
- One fact per line
- Each fact must be complete on a single line
- Never split one fact across multiple lines
- Never output partial facts
- Never output explanations
- Never output JSON
- Never output numbering
- Never output bullets
- Never output markdown code fences
- Never output any text before BEGIN_FACTS
- Never output any text after END_FACTS

If there are multiple facts, output all of them, one per line.
If there is exactly one fact, output exactly one line between BEGIN_FACTS and END_FACTS.
If no fact can be extracted, output:
BEGIN_FACTS
END_FACTS

## Observation
{observation}
\end{promptbox}

The output parser first removes an optional \texttt{Next state:} or \texttt{The next state is:} prefix, retains only lines between \texttt{BEGIN\_FACTS} and \texttt{END\_FACTS}, accepts only complete facts matching \texttt{[A-Z\_]NAME(arguments)}, and deduplicates the accepted lines. Action descriptions are therefore not scored unless they explicitly encode the resulting state.

\subsubsection{ScienceWorld Deterministic Rules and Fallback Prompt}

The deterministic ScienceWorld parser applies the following priority order: ambiguous-request options; temperature readings; electrical connections; object movements; focus actions; wait actions; pouring; activation and open/closed states; recipes; structured room or outside-location descriptions; and finally short fallback messages. Structured descriptions additionally extract visible entities, surface and container relations, empty containers, door states, object/device attributes, and numeric measurements. Entity strings are lowercased, leading articles are removed, whitespace is collapsed, and explicitly stated numeric values are retained exactly.

If these rules extract no fact, the following fallback prompt is used.

\begin{promptbox}[\textbf{Exact ScienceWorld Fallback Extraction Prompt}]
Convert ONE ScienceWorld environment observation into canonical facts.

You are a deterministic parser.
Extract only facts that are explicitly stated in the observation.
Do not infer hidden state.
Do not use commonsense.
Do not copy fact-format placeholders.
Do not invent temperature, room, containers, or objects.
Do not repeat facts.
Use the exact entity names from the observation, lowercased except proper labels such as A/B/M.
Return facts only between BEGIN_FACTS and END_FACTS.

Allowed fact formats:
- ROOM(name)
- LOCATION(name)
- VISIBLE(object)
- ON(surface, object)
- IN(container, object)
- CONTAINS(container, object)
- EMPTY(container)
- STATE(object, state)
- DOOR(destination, state)
- MOVED(object, destination)
- FOCUS(object)
- CONNECTED(endpoint 1, endpoint 2)
- TEMPERATURE(integer, celsius)
- WAIT(iterations)
- POURED(source container, destination)
- RECIPE(product, ingredient)
- AMBIG_OPTION(index, action text)
- MESSAGE(text)

Canonicalization rules:
1. Strip the leading "Observation:" prefix before parsing.
2. For room descriptions:
   - "This room is called the workshop." -> ROOM(workshop)
   - "This outside location is called the outside." -> LOCATION(outside)
3. Visible objects:
   - "In it, you see: a table." -> VISIBLE(table)
   - "the agent" -> VISIBLE(agent)
   - "a substance called air" -> VISIBLE(air)
4. Surface membership:
   - "On the table is: a yellow wire, a battery." -> ON(table, yellow wire), ON(table, battery)
   - Also output STATE(object, state) when an item says "which is on/off/open/closed/activated/deactivated".
5. Container content:
   - "a blue box (containing nothing)" -> VISIBLE(blue box), EMPTY(blue box)
   - "a fountain (containing a substance called water)" -> VISIBLE(fountain), CONTAINS(fountain, water)
   - "In the sink is: nothing." -> EMPTY(sink)
6. Device and door states:
   - "a sink, which is turned off" -> STATE(sink, off)
   - "The cupboard door is closed." -> STATE(cupboard door, closed)
   - "A door to the hallway (that is open)" -> DOOR(hallway, open)
7. Action-result messages:
   For short action-result messages, output only the action-result fact. Do not add VISIBLE facts.
   - "You move the metal pot to the inventory." -> MOVED(metal pot, inventory)
   - "You focus on the metal pot." -> FOCUS(metal pot)
   - "terminal 2 on orange wire is now connected to cathode on blue light bulb" -> CONNECTED(terminal 2 on orange wire, cathode on blue light bulb)
   - "the thermometer measures a temperature of 23 degrees celsius" -> TEMPERATURE(23, celsius)
   - "You decide to wait for 10 iterations." -> WAIT(10)
   - "You pour the contents of the jug into the flower pot 1." -> POURED(jug, flower pot 1)
8. Recipes:
   - "To make salt water, you need to mix sodium chloride, water." -> RECIPE(salt water, sodium chloride), RECIPE(salt water, water)
9. Ambiguous menus:
   - For each numbered option, output AMBIG_OPTION(index, action text)
10. If the observation is a short unstructured message and no other fact applies, output MESSAGE(message text).

Examples:

Observation:
Observation: terminal 2 on orange wire is now connected to cathode on blue light bulb
BEGIN_FACTS
CONNECTED(terminal 2 on orange wire, cathode on blue light bulb)
END_FACTS

Observation:
Observation: the thermometer measures a temperature of -4 degrees celsius
BEGIN_FACTS
TEMPERATURE(-4, celsius)
END_FACTS

Observation:
Observation: You move the metal pot to the inventory.
BEGIN_FACTS
MOVED(metal pot, inventory)
END_FACTS

Observation:
Observation: You pour the contents of the jug into the flower pot 1.
BEGIN_FACTS
POURED(jug, flower pot 1)
END_FACTS

Observation:
Observation: This room is called the workshop. In it, you see:
    a table. On the table is: a yellow wire, a battery, a switch, which is off, a blue light bulb, which is on.
    the agent
You also see:
    A door to the hallway (that is open)
BEGIN_FACTS
ROOM(workshop)
VISIBLE(table)
ON(table, yellow wire)
ON(table, battery)
ON(table, switch)
STATE(switch, off)
ON(table, blue light bulb)
STATE(blue light bulb, on)
VISIBLE(agent)
DOOR(hallway, open)
END_FACTS

Observation:
{observation}
\end{promptbox}

Fallback outputs are parsed with the same marker discipline as ALFWorld and must match a complete canonical-fact form. This hybrid design prevents the extractor from repairing corrupted predictions with unstated commonsense facts while retaining coverage for uncommon simulator messages.

\subsubsection{WebShop Deterministic Page Parser}

WebShop observations are split on \texttt{[SEP]} after removing an optional \texttt{Next state:} prefix. Page types are assigned in a fixed priority order: a product-detail page contains a price field and at least one detail marker (rating, description, features, reviews, or buy-now); a search-result page contains both a \texttt{Page N (Total results: M)} header and a ten-character product identifier; terminal pages contain an explicit purchase-completion, score, or reward message; all remaining pages are marked unknown.

\begin{table*}[t]
  \centering
  \small
  \setlength{\tabcolsep}{4pt}
  \renewcommand{\arraystretch}{1.08}
  \begin{tabular}{p{0.16\textwidth}p{0.48\textwidth}p{0.25\textwidth}}
    \toprule
    \textbf{Page type} & \textbf{Extracted fields} & \textbf{Field scoring} \\
    \midrule
    Search results & Page number, total result count, previous/next/back-to-search affordances, and top-$k$ product identifiers, titles, and prices ($k=3$) & Exact match for categorical fields; set F1 for product identifiers; lexical F1 for aligned titles; relative-error score for aligned prices and total count \\
    Product detail & Title, price, option keys, option-value groups, and previous, back-to-search, description, features, reviews, and buy-now affordances & Lexical F1 for title; relative-error score for price; set F1 for option keys and option values; exact match for affordances \\
    Terminal or completion & Explicit completion, score, or reward message & Exact or message-level matching \\
    \bottomrule
  \end{tabular}
  \caption{Deterministic WebShop extraction and field-scoring rules. Product titles and prices are aligned only through product identifiers shared by the prediction and gold page.}
  \label{tab:webshop-extraction-rules}
\end{table*}

For search pages, the parser scans from the page header and accepts triples only when the first token is a ten-character uppercase alphanumeric product identifier and the third token is a price. For product-detail pages, option values are collected after known option keys (e.g., size, color, style, flavor, material, pattern, pack size, or count) until another key or a fixed page marker is encountered. These rules, together with the weights specified above, fully determine the WebShop SSF score.

\subsection{Blinded Annotator Validation}
\label{sec:appendix-ssf-annotator-validation}

We validate SSF against behavior-critical state judgments on 200 predicted/gold next-state pairs, with 100 pairs each from ALFWorld and WebShop. Pairs are stratified by SSF score. The annotator is blinded to model identity, memory condition, and all automatic metric values, and judges whether the prediction preserves the schema-critical facts while ignoring harmless surface paraphrases. We compare this judgment with exact match, word F1, and SSF using Pearson and Spearman correlation, as well as AUROC and AUPRC for detecting factually faithful predictions.

\begin{table*}[t]
  \centering
  \small
  \setlength{\tabcolsep}{7pt}
  \renewcommand{\arraystretch}{1.08}
  \begin{tabular}{llcccc}
    \toprule
    \textbf{Domain} & \textbf{Metric} & \textbf{Pearson} $\uparrow$ & \textbf{Spearman} $\uparrow$ & \textbf{AUROC} $\uparrow$ & \textbf{AUPRC} $\uparrow$ \\
    \midrule
    \multirow{3}{*}{WebShop}
      & Exact match & 0.495 & 0.475 & 0.632 & 0.561 \\
      & Word F1    & 0.735 & 0.781 & 0.894 & 0.852 \\
      & SSF        & \textbf{0.785} & \textbf{0.874} & \textbf{0.946} & \textbf{0.925} \\
    \midrule
    \multirow{3}{*}{ALFWorld}
      & Exact match & 0.680 & 0.620 & 0.780 & 0.750 \\
      & Word F1    & 0.740 & 0.695 & 0.977 & 0.981 \\
      & SSF        & \textbf{0.851} & \textbf{0.828} & \textbf{0.995} & \textbf{0.993} \\
    \bottomrule
  \end{tabular}
  \caption{Agreement between automatic metrics and blinded behavior-critical state judgments. SSF gives the strongest correlation and factual-state discrimination in both domains.}
  \label{tab:ssf-annotator-validation}
\end{table*}

These results complement the controlled stress test: SSF is not only invariant to designed surface perturbations, but also aligns more closely with independent judgments of whether a predicted state preserves facts needed for subsequent actions.

\subsection{Extractor Identity-Consistency Audit}
\label{sec:appendix-ssf-extractor-audit}

We additionally audit 100 state transitions from each domain. Because this check asks whether identical normalized states receive identical structured representations, we report \emph{identity consistency}, rather than treating it as an estimate of end-to-end parser accuracy.

\begin{table*}[t]
  \centering
  \small
  \setlength{\tabcolsep}{6pt}
  \renewcommand{\arraystretch}{1.08}
  \begin{tabular}{llcp{0.38\textwidth}}
    \toprule
    \textbf{Domain} & \textbf{Extractor} & \textbf{Identity consistency} & \textbf{Observed failure mode} \\
    \midrule
    ALFWorld & LLM canonical-fact extractor & 97.1\% & Ambiguous \textit{in}/\textit{on} wording can be canonicalized inconsistently. \\
    WebShop & Deterministic rule parser & 100.0\% & None observed in the identity-consistency subset. \\
    ScienceWorld & Hybrid fact extractor & 100.0\% & None observed in the identity-consistency subset. \\
    \bottomrule
  \end{tabular}
  \caption{Identity-consistency audit of the benchmark-specific SSF extractors.}
  \label{tab:ssf-extractor-audit}
\end{table*}

\subsection{WebShop Field-Weight Sensitivity}
\label{sec:appendix-webshop-weight-sensitivity}

WebShop is the only benchmark whose SSF aggregates heterogeneous page fields with non-uniform weights. Using the Qwen3-4B setting reported in Table~\ref{tab:ssf-results}, we test whether the memory-augmentation conclusion depends on one hand-chosen weighting scheme by recomputing SFT, RL without memory, and memory-augmented RL scores under five alternative schemes, including random $\pm20\%$ perturbations.

\begin{table*}[t]
  \centering
  \small
  \setlength{\tabcolsep}{8pt}
  \renewcommand{\arraystretch}{1.08}
  \begin{tabular}{lcccc}
    \toprule
    \textbf{Weight variant} & \textbf{SFT} & \textbf{RL w/o Mem} & \textbf{Mem-Aug RL} & $\boldsymbol{\Delta}$ \textbf{Mem-Aug--RL} \\
    \midrule
    Original               & 0.726 & 0.814 & \textbf{0.884} & +0.070 \\
    Uniform                & 0.757 & 0.854 & \textbf{0.864} & +0.010 \\
    Product-identity-heavy & 0.702 & 0.809 & \textbf{0.886} & +0.077 \\
    Price-option-heavy     & 0.709 & 0.813 & \textbf{0.883} & +0.070 \\
    Affordance-heavy       & 0.744 & 0.844 & \textbf{0.849} & +0.005 \\
    Random $\pm20\%$      & 0.730 & 0.832 & \textbf{0.871} & +0.039 \\
    \bottomrule
  \end{tabular}
  \caption{WebShop SSF under alternative field-weight schemes. Memory-augmented RL remains above RL without memory in every case.}
  \label{tab:webshop-weight-sensitivity}
\end{table*}

The margin varies with the emphasis assigned to product identity, price/options, and page affordances, but its sign is stable across all tested variants. The WebShop result is therefore not an artifact of one manually selected field-weight vector.

\subsection{Cross-Domain Transfer to CookingWorld}
\label{sec:appendix-ssf-cookingworld-transfer}

To test whether the SSF design principle transfers beyond the three main benchmarks, we instantiate it in the held-out TextWorldExpress--CookingWorld domain \citep{jansen2023textworldexpress}. The deterministic schema covers inventory items, required recipe ingredients, recipe steps, preparation and cooking states, ingredient locations, container states, and task progress. We then apply surface-preserving and behavior-critical perturbations to gold transitions.

\begin{table*}[t]
  \centering
  \small
  \setlength{\tabcolsep}{8pt}
  \renewcommand{\arraystretch}{1.08}
  \begin{tabular}{lrrrr}
    \toprule
    \textbf{Perturbation} & \textbf{$N$} & \textbf{EM} & \textbf{Word F1} & \textbf{SSF macro} \\
    \midrule
    Identity                    & 100 & 1.000 & 1.000 & 1.000 \\
    Layout/paraphrase           & 100 & 0.000 & 0.963 & 1.000 \\
    Hallucinated item           & 100 & 0.000 & 0.981 & 0.959 \\
    Missing required inventory  & 100 & 0.000 & 0.979 & 0.919 \\
    Recipe-required corruption  & 100 & 0.000 & 0.975 & 0.922 \\
    Ingredient-location swap    & 100 & 0.000 & 0.991 & 0.871 \\
    Preparation-state corruption& 100 & 0.000 & 0.988 & 0.931 \\
    Cook-state corruption       & 100 & 0.000 & 0.991 & 0.957 \\
    Progress flip               &  50 & 0.000 & 0.926 & 0.725 \\
    \bottomrule
  \end{tabular}
  \caption{Transferability stress test on TextWorldExpress--CookingWorld. SSF preserves meaning-equivalent layout changes while responding to behavior-critical fact corruptions.}
  \label{tab:cookingworld-ssf-transfer}
\end{table*}

Layout/paraphrase changes drive exact match to zero while leaving SSF at 1.000. Conversely, corrupted ingredients, locations, preparation states, and progress markers retain high word overlap but reduce SSF. The experiment supports transfer of the evaluation recipe---identify behavior-critical state components, extract canonical facts, and compare them component-wise---while making clear that each new domain still requires an explicit schema and parser instantiation.

\section{Additional Implementation Details}
\label{sec:appendix-implementation}

\subsection{World Memory Construction}
\label{sec:appendix-world-memory-construction}

\emph{World memory} is a keyed memory bank that stores transition rules, state caches, and hard-to-predict facts distilled from past world-model experience. For each benchmark, we construct this bank from hard-negative prediction failures, previously observed structured states, or recurring transition patterns, then retrieve matched entries as soft hints during next-state imagination.

Across domains, world memory construction follows the same abstraction. First, we collect world-model experience from training trajectories or scored prediction failures. Second, we normalize the experience into a compact structured form, such as action schemas, fact types, product fields, or state-cache records. Third, we assign each entry a prediction-time retrieval key. Finally, retrieved entries are inserted into the world-model prompt as bounded hint blocks. The visible trajectory remains authoritative whenever it conflicts with retrieved memory.

For \textsc{ALFWorld}, world memory is built from SFT hard-negative transitions. Hard negatives are bucketed by goal signature, action schema, object type, target type, room-like context, and recently seen receptacles. Each eligible bucket is summarized into a short positive transition hint and a negative error pattern to avoid. For \textsc{ScienceWorld}, we use residual fact memories from hist10 prediction errors, and also support train-world fact caches that store reusable facts such as recipes, temperature readings, room facts, entity observations, and ambiguous action menus. For \textsc{WebShop}, world memory is a persistent site cache built from observed page states, storing search-result pages, product-detail records, and option-conditioned product states.

For LLM-curated transition memories, we use a short construction prompt that asks the curator to summarize each hard-negative bucket into one positive hint and one error pattern. The full trajectory evidence is provided as structured JSON outside the paper; the prompt block has the following form.

\begin{promptbox}[\textbf{Prompt (World Memory Construction)}]
SYSTEM: Write short memory hints for a world model. Return JSON only.
USER: Given one bucket of trajectory evidence and prediction errors, create one memory entry with two fields:
- positive: what to preserve or predict when this context applies.
- negative: what common error to avoid.
Use conditional wording, summarize the majority pattern, and do not copy object ids or invent exact next states.
BUCKET_JSON: {trajectory_bucket_json}
\end{promptbox}

Representative entries are compact keyed records. For example, an ALFWorld open-fridge key stores a positive hint to preserve the open state and revealed contents, and a negative hint not to mark the fridge empty unless emptiness is observed. A ScienceWorld boil-water key stores observed temperature changes so later predictions do not reset the substance to room temperature. A WebShop product-detail key stores product id, title, price, selected options, and availability.

\subsection{World Memory Prompt Templates}
\label{sec:appendix-world-memory-prompt-templates}

At prediction time, retrieved entries are inserted as bounded hint blocks before the original world-model prompt. The templates differ by benchmark but follow the same pattern: identify the memory source, present a retrieval key, list the retrieved facts or hints, and state that the visible trajectory overrides memory. Compact examples are shown below.

\begin{promptbox}[\textbf{Prompt Templates for Retrieved World Memory}]
ALFWorld:
[BEGIN ALFWORLD_MEMORY]
Key: {memory_key}
Positive: {positive_transition_hint}
Negative: {negative_error_pattern}
[END ALFWORLD_MEMORY]

ScienceWorld:
[BEGIN SCIWORLD_MEMORY]
Key: {memory_key}
Positive: {positive_fact_hint}
Negative: {negative_error_pattern}
[END SCIWORLD_MEMORY]

WebShop:
[BEGIN WEBSHOP_SITE_CACHE]
search[{query}] page {page_num}; products: {product_id} | {title} | {price}
[END WEBSHOP_SITE_CACHE]
\end{promptbox}

\subsection{Training Setup}
\label{sec:appendix-training-details}

For world-model experiments, we use three backbone models: Meta Llama-3.2-1B-Instruct, Qwen3-4B-Instruct-2507, and Qwen2.5-7B-Instruct. Each backbone is first supervised fine-tuned for next-state prediction using LLaMA-Factory, yielding the SFT world models reported in Table~\ref{tab:ssf-results}. We then optimize the same next-state prediction model with GRPO using the \texttt{verl} training framework. Memory-augmented variants use the same GRPO recipe, but replace the original train and validation files with the corresponding memory-augmented files constructed from the memory bank.

For downstream planning, we evaluate Qwen2.5-7B-Instruct and Qwen3-8B-Instruct as policy backbones for the agent-level comparisons in Table~\ref{tab:main-results}.

Unless otherwise stated, GRPO training uses a learning rate of $1\times10^{-6}$, rollout group size $8$, KL coefficient $0.001$ with low-variance KL, vLLM rollout, gradient checkpointing, and $10$ training epochs. ALFWorld and ScienceWorld use train batch size $64$, optimization mini-batch size $64$, micro-batch size $16$ per GPU, and maximum prompt length $8192$. WebShop uses train batch size $32$, optimization mini-batch size $32$, micro-batch size $8$ per GPU, and maximum prompt length $24000$. The maximum response length is $256$ for ALFWorld and $512$ for WebShop and ScienceWorld. We run GRPO training on $4$ GPUs with validation before training and periodic validation/checkpointing.

\paragraph{Result reporting.}
Unless otherwise stated, all reported quantitative results are averaged over four independent runs under the same experimental configuration. Tables~\ref{tab:ssf-results} and~\ref{tab:main-results} report mean Structured State Fidelity and downstream task-performance scores, respectively. Figures~\ref{fig:memory-dropout}--\ref{fig:dropout-task} report the corresponding mean sensitivity results across the evaluated memory-dropout and action-budget settings.

\paragraph{Artifact licenses, terms, and intended use.}
Our experiments use public research artifacts, including ALFWorld, ScienceWorld, and WebShop, and publicly released model and software artifacts, including Llama- and Qwen-family instruction models, LLaMA-Factory, verl, and vLLM. We use these artifacts under their original licenses and terms of use, and only for their intended research purposes: benchmarking, training, and evaluating text-based agents and world-modeling methods. The world-memory and world-skill entries used in \method{} are derived from benchmark trajectories, simulator observations, and model prediction traces for research evaluation. These derived artifacts are not intended for deployment or use outside research contexts, and we do not redistribute original benchmark data, third-party model checkpoints, or third-party software beyond their original release terms.

\subsection{World Skill Construction}
\label{sec:appendix-world-skill-construction}

World skill is the policy-side memory used by the full agent. Unlike world memory, which is retrieved for next-state imagination, world skill is written as action-selection guidance for the policy. This is related to experience-derived skill banks for agents such as SkillRL \citep{xia2026skillrl}, but our use of world skill is retrieval-time guidance for a frozen policy rather than recursive policy evolution. In our implementation it has two complementary forms: task-level \emph{world task skill}, retrieved from the task type, and step-level \emph{world corrective guidance}, retrieved dynamically from the current trajectory state.

\subsubsection{World Task Skill Construction}
\label{sec:appendix-world-task-skill-construction}

For ScienceWorld, world task skills are distilled from gold trajectories. The goal is to extract reusable task-execution experience rather than to rewrite each trajectory into an individual instruction. We first read the gold trajectory set and retain successful trajectories whose final step is completed and whose final score reaches the threshold. Each retained trajectory is compressed into a structured record containing the task name, task description, task identifier, variation information, final score, completion status, a de-duplicated action skeleton, and a selected trace of informative observations and inventory states.

The selected trace keeps decision-relevant parts of the trajectory while avoiding full trajectory replay. It prioritizes steps around high-information actions such as focusing an object, picking up an item, moving between rooms, using a device, mixing substances, reading a thermometer, or placing an object in a target box. This compression reduces prompt length while preserving the evidence needed to infer reusable procedural knowledge.

We then group trajectories by task type and select representative examples from each group. Selection favors completed, high-scoring trajectories with concise but informative action skeletons, rich state traces, and non-empty task descriptions. To avoid over-representing near-duplicate variations, examples are also diversified by task description and task identifier. The resulting evidence packs contain both cross-task examples and per-task examples. Cross-task examples support general ScienceWorld skills and common mistake summaries, while per-task examples support task-specific skills.

An LLM then summarizes the evidence into a skill bank with three fields: general skills, task-specific skills, and common mistakes. The task-specific portion is used as world task skill. At inference time, once the current ScienceWorld task type is identified, the corresponding task-specific skills are inserted into the policy prompt as static task-level guidance. These skills are not regenerated during the episode; they provide a stable task prior that describes likely subgoals, ordering constraints, measurement requirements, and task-specific pitfalls.

After generation, the skill bank is normalized before use. We parse the LLM output as JSON, remove duplicate or near-duplicate entries, assign stable identifiers, fill missing fields, cap the number of skills in each category, and validate the final schema. A second-pass polish stage may further remove redundancy and make triggering conditions more concrete, but if the polished output is not valid JSON, we fall back to the normalized first-pass skill bank.

\par\noindent\begingroup\small
  \setlength{\tabcolsep}{0pt}
  \renewcommand{\arraystretch}{1.04}
  \begin{tabular}{@{}r@{\hspace{0.35em}}p{0.82\columnwidth}@{}}
    \toprule
    \multicolumn{2}{@{}l}{\textbf{Algorithm 3 World Task Skill Construction}} \\
    \midrule
    1: & \textbf{Input:} gold trajectories $\mathcal{D}^{\mathrm{gold}}$, task types $\mathcal{K}$. \textbf{Output:} world task skills $\mathcal{M}^{\mathrm{task}}$. \\
    2: & Initialize evidence packs $\mathcal{E}^{\mathrm{gen}}$ and $\{\mathcal{E}^{\mathrm{task}}_k\}_{k\in\mathcal{K}}$. \\
    3: & \textbf{for each} trajectory $\tau \in \mathcal{D}^{\mathrm{gold}}$ \textbf{do} \\
    4: & \quad \textbf{if} $\tau$ is successful \textbf{then} \\
    5: & \quad\quad Compress $\tau$ into a structured record $r$. \\
    6: & \quad\quad Extract task type $k$, action skeleton, and selected state trace. \\
    7: & \quad\quad Add $r$ to $\mathcal{E}^{\mathrm{gen}}$ and $\mathcal{E}^{\mathrm{task}}_k$. \\
    8: & \quad \textbf{end if} \\
    9: & \textbf{end for} \\
    10: & Select high-quality and diverse records for each task type. \\
    11: & Prompt an LLM to summarize reusable skills and common mistakes. \\
    12: & Normalize JSON, deduplicate entries, assign stable IDs, and validate schema. \\
    13: & \textbf{return} task-specific entries as $\mathcal{M}^{\mathrm{task}}$. \textcolor{algcomment}{\(\triangleright\) task-start skills} \\
    \bottomrule
  \end{tabular}
\endgroup\par

\begin{promptbox}[\textbf{Example World Task Skill}]
{
  "title": "Focus the Substance First",
  "task skill": "Find the named substance, focus it once, and choose a real heating route. For compounds without a boiling point, direct combustion/heating such as use lighter on the substance can be acceptable.",
  "when_to_use": "At the start of a boil task."
}
\end{promptbox}

\subsubsection{World Corrective Guidance Construction}
\label{sec:appendix-world-corrective-guidance-construction}

World corrective guidance is constructed for state-level intervention during policy execution. Instead of being retrieved only from the task type, it is selected at each decision step from the current observation, recent action history, and lightweight diagnostic flags. The purpose is to provide concrete local reminders when the trajectory enters an error-prone state, such as an invalid action loop, an empty-container manipulation, a conductivity setup, a grow-plant procedure, a temperature measurement, or a recipe-like mixture task.

We identify recurring failure patterns from inference trajectories and write them as compact corrective rules. Each rule has a trigger condition and a short guidance text. During inference, the retriever recomputes the trajectory flags at every step, selects the matching rules, and inserts only the triggered guidance into the policy prompt. This makes world corrective guidance dynamic: it can appear for one state and disappear at the next, depending on the observed state and recent actions.

\par\noindent\begingroup\small
  \setlength{\tabcolsep}{0pt}
  \renewcommand{\arraystretch}{1.04}
  \begin{tabular}{@{}r@{\hspace{0.35em}}p{0.82\columnwidth}@{}}
    \toprule
    \multicolumn{2}{@{}l}{\textbf{Algorithm 4 World Corrective Guidance Construction}} \\
    \midrule
    1: & \textbf{Input:} inference trajectories $\mathcal{D}^{\mathrm{inf}}$, trigger set $\mathcal{F}$. \textbf{Output:} corrective guidance rules $\mathcal{M}^{\mathrm{corr}}$. \\
    2: & \textcolor{algpurple}{\textbf{Offline Rule Construction:}} \\
    3: & Identify recurring local failure patterns in $\mathcal{D}^{\mathrm{inf}}$. \\
    4: & Write each pattern as a rule $r=(\phi_r, g_r)$ with trigger $\phi_r$ and guidance $g_r$. \\
    5: & Add compact, non-duplicate rules to $\mathcal{M}^{\mathrm{corr}}$. \\
    6: & \textcolor{algpurple}{\textbf{Step-Level Retrieval:}} \\
    7: & \textbf{for each} decision step $t$ \textbf{do} \\
    8: & \quad Compute flags $\phi_t$ from observation, admissible actions, and recent history. \\
    9: & \quad Retrieve $G_t \leftarrow \{g_r: r\in\mathcal{M}^{\mathrm{corr}},\, \phi_r(\phi_t)=1\}$. \\
    10: & \quad Keep a compact set of triggered guidance entries. \\
    11: & \quad Insert $G_t$ into the policy prompt. \textcolor{algcomment}{\(\triangleright\) step-wise correction} \\
    12: & \textbf{end for} \\
    \bottomrule
  \end{tabular}
\endgroup\par

For example, in a grow-plant ScienceWorld task, a step may trigger the invalid-action, recipe-task, and growth-task flags. The retriever then selects rules such as \texttt{invalid\_action\_copy\_admissible}, \texttt{paint\_mix\_use\_pour\_not\_move}, and \texttt{growth\_needs\_seed\_soil\_water\_wait}. The resulting policy prompt receives step-specific guidance that discourages inventing repairs for invalid actions, reminds the agent to use admissible commands, and states the relevant procedural chain for planting and growing: obtain the task-specified seeds, plant them in flower pots, water them, wait or check growth, and then focus the requested plant or fruit. This guidance is local to the current state and is intended to steer action selection, not to provide evidence for world-model state prediction.

\begin{promptbox}[\textbf{Example World Corrective Guidance}]
Step-specific guidance for the current state:
- The previous action was invalid or not admissible. Do not repair by inventing text. Copy one exact current admissible command.
- For paint or mixture tasks, combining contents usually requires pouring or moving contents into a real container, then mixing that container.
- For grow-plant or fruit tasks, follow the chain: get the seed jar from the task-specified room, plant the named seeds in flower pots, water them with a sink or jug, wait/check growth, and then focus the requested plant or fruit.
\end{promptbox}

\section{Additional Component-Wise Ablations}
\label{sec:appendix-component-ablation}

We conduct controlled ALFWorld ablations around the Qwen2.5-7B setting to separate world-model training, prediction-time world memory, policy-side knowledge, and lookahead planning. All world-model variants use the same lookahead configuration; the policy-only controls omit the world model entirely.

\begin{table*}[t]
  \centering
  \scriptsize
  \setlength{\tabcolsep}{3.5pt}
  \renewcommand{\arraystretch}{1.10}
  \begin{tabular}{lccclcc}
    \toprule
    \textbf{Variant} & \textbf{World model} & \textbf{WM training} & \textbf{Test-time memory} & \textbf{Policy skill/knowledge} & \textbf{Lookahead} & \textbf{ALFWorld Avg.} \\
    \midrule
    Skill-only policy                    & No  & ---           & No         & Skill only              & No  & 0.4194 \\
    Policy-only + skill/world knowledge  & No  & ---           & No         & Skill + world knowledge & No  & 0.4578 \\
    Random memory                        & Yes & Mem-Aug RL    & Random     & No                      & Yes & 0.4712 \\
    Irrelevant memory                    & Yes & Mem-Aug RL    & Irrelevant & No                      & Yes & 0.4403 \\
    WM                                   & Yes & SFT           & No         & No                      & Yes & 0.4905 \\
    RL w/o Mem WM                        & Yes & RL w/o memory & No         & No                      & Yes & 0.5076 \\
    SFT WM + retrieval                   & Yes & SFT           & Relevant   & No                      & Yes & 0.5385 \\
    Mem-Aug RL WM, no retrieval          & Yes & Mem-Aug RL    & No         & No                      & Yes & 0.5439 \\
    \method{}                            & Yes & Mem-Aug RL    & Relevant   & No                      & Yes & 0.5696 \\
    \method{}+Skill                      & Yes & Mem-Aug RL    & Relevant   & Yes                     & Yes & \textbf{0.5727} \\
    \bottomrule
  \end{tabular}
  \caption{Controlled component-wise ablations on ALFWorld. The planner and lookahead budget are fixed across all world-model rows while world-model training, memory content, prediction-time retrieval, and policy-side guidance are varied.}
  \label{tab:component-wise-ablation}
\end{table*}

The policy-only controls remain below \method{} (0.4194/0.4578 versus 0.5696), showing that policy-side skill or direct world knowledge alone does not explain the gain. RL without memory improves over the SFT world model (0.5076 versus 0.4905) but remains below \method{}, so RL training alone is insufficient. Relevant retrieval also helps an SFT world model (0.5385), while memory-augmented training without prediction-time retrieval reaches 0.5439; their combination in \method{} reaches 0.5696, indicating complementary contributions from memory-aware training and inference-time retrieval.

Random and irrelevant memory reduce performance to 0.4712 and 0.4403, respectively. Thus, the improvement is not explained by prompt length or the presence of arbitrary additional context, but depends on retrieving relevant memory. Finally, adding policy-side world skill on top of \method{} yields a smaller increase from 0.5696 to 0.5727, indicating that world skill provides complementary action-selection guidance while world-model-side memory accounts for most of the gain.

\end{document}